\documentclass[11pt]{article}

\usepackage[preprint]{acl}

\usepackage{times}
\usepackage{latexsym}

\usepackage[T1]{fontenc}
\usepackage[utf8]{inputenc}

\usepackage{microtype}

\usepackage{inconsolata}

\usepackage{graphicx}
\usepackage{todonotes}

\usepackage{xspace}
\usepackage{booktabs}
\usepackage{amsmath}
\usepackage{multirow}
\usepackage{colortbl}

\newcommand{\dsetname}{\textsc{WikiVideo}\xspace}
\newcommand{\methodname} {\textsc{CAG}\xspace}
\newcommand{\concatg}{\textsc{Concat-0}\xspace}
\newcommand{\concatr}{\textsc{Concat-RePrompt}\xspace}
\newcommand{\concatrs}{\textsc{Concat-R}\xspace}
\newcommand{\summg}{\textsc{CAG-0}\xspace}
\newcommand{\cagr}{\textsc{CAG-R}\xspace}

\title{WikiVideo: Article Generation from Multiple Videos}

\author{
Alexander Martin\textsuperscript{\rm 1} 
\quad Reno Kriz\textsuperscript{\rm 1,2*}
\quad William Walden\textsuperscript{\rm 1,2*}
\quad Kate Sanders\textsuperscript{\rm 1} \\
\textbf{Hannah Recknor}\textsuperscript{\rm 1,2}
\quad \textbf{Eugene Yang}\textsuperscript{\rm 1} 
\quad \textbf{Francis Ferraro}\textsuperscript{\rm 3} 
\quad \textbf{Benjamin Van Durme}\textsuperscript{\rm 1,2} 
\\
  \textsuperscript{1}Johns Hopkins University\quad \textsuperscript{2}Human Language Technology Center of Excellence\quad \\
  \textsuperscript{3}University of Maryland Baltimore County 
  \\
  \texttt{\small{\{amart233, vandurme\}@jhu.edu}}
}

\begin{document}
\maketitle
\begin{abstract}

We introduce the task of grounded article generation with the goal of creating a Wikipedia-style article from multiple diverse videos about real-world events---from natural disasters to political elections---where all the information in the article is supported by video evidence. Videos are intuitive sources for retrieval-augmented generation (RAG), but most contemporary RAG workflows focus heavily on text while existing methods for video-based summarization focus on low-level scene understanding rather than high-level event semantics. To close this gap, we introduce \dsetname, a benchmark consisting of expert-written articles and densely annotated videos that provide evidence for articles' claims, facilitating the integration of video into RAG pipelines and enabling the creation of in-depth content that is grounded in multimodal sources. We further propose \textbf{C}ollaborative \textbf{A}rticle \textbf{G}eneration (\methodname), a novel interactive method for article creation from multiple videos. \methodname leverages an iterative interaction between an r1-style reasoning model and a VideoLLM to draw higher-level inferences about the target event than is possible with VideoLLMs alone, which fixate on low-level visual features. We benchmark state-of-the-art VideoLLMs and \methodname in both oracle retrieval and RAG settings and find that \methodname consistently outperforms alternative methods, while 
suggesting intriguing 
avenues for future work.\footnote{Data and code can be found here: \url{https://github.com/alexmartin1722/wikivideo}}

\end{abstract}

\section{Introduction}
\label{sec:intro}
\begin{figure*}[t]
    \centering
    \includegraphics[width=\linewidth]{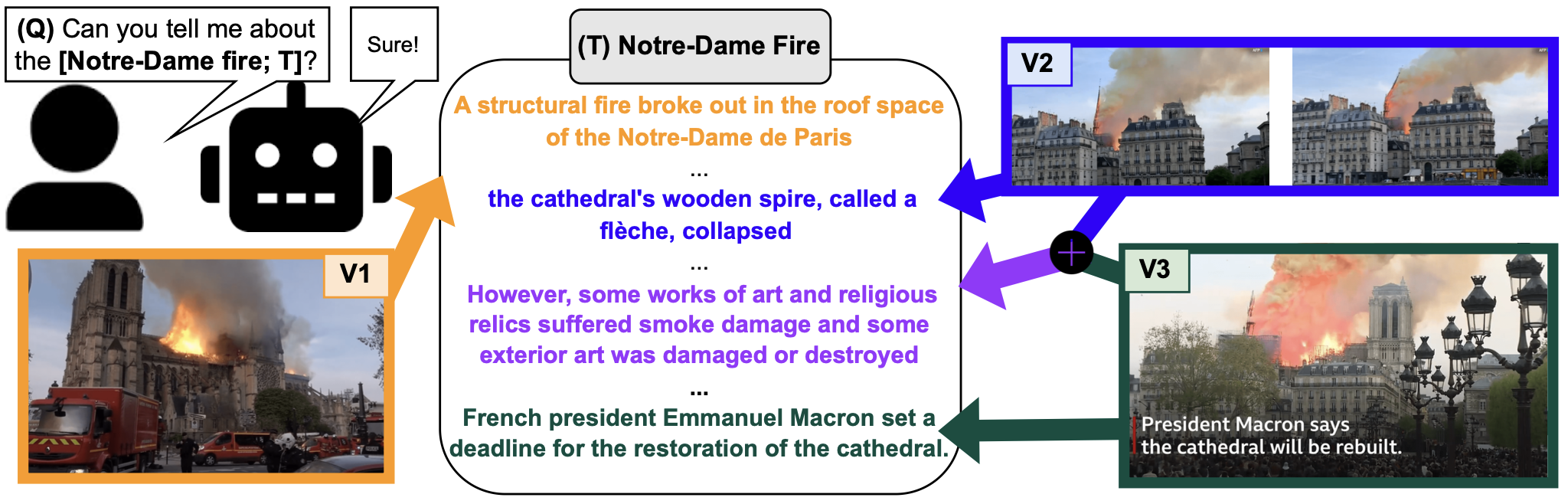}
    \caption{\dsetname introduces the task of Article Generation from Multiple Videos, which requires writing a high-level article in the style of a Wikipedia lead, given a target event ($T$), a query about that event ($Q$), and a collection of $Q$-relevant videos ($V$). All claims in the article are grounded in visual, audio, and/or OCR content of video(s) in $V$ (indicated by matching colors between text and frame borders above).}
    \label{fig:task}
    \vspace{-1em}
\end{figure*}
Audiovisual media is becoming an increasingly dominant form of online information consumption. From firsthand, ``in the wild'' video footage of natural disasters to professionally edited news coverage of major political events, videos serve as rich sources of information for producing factual, grounded articles. Especially for \emph{actively unfolding} events, grounding articles in video not only can potentially combat misinformation among readers, but can also provide a useful tool for journalists and other writers to quickly synthesize information about new developments. \autoref{fig:task} motivates this task, taking an information request and producing a Wikipedia-style article with references that ground the article in the supporting video content.

Current methods and resources for article generation overwhelmingly rely on textual sources~\citep[][\emph{i.a.}]{liu2018generating, barham2023megawikamillionsreportssources, lawrie2024overviewtrec2023neuclir, shao2024assistingwritingwikipedialikearticles}, while video understanding benchmarks largely focus on low-level tasks such as entity-centric question answering or captioning \citep{7780940, Krishna_2017_ICCV, zhou2019activitynetqa, Li_2022_CVPR, lin2024videoxum}. \citet{kriz2024multivent20massivemultilingual} show that models trained on such tasks often fail in more realistic settings that require understanding of high-level semantics, (e.g.) as conveyed in articles about actual events. The MultiVENT benchmarks \citep{sanders2023multivent,kriz2024multivent20massivemultilingual} are distinctive in focusing on \emph{major real-world events} as depicted in \emph{multiple} videos---spanning firsthand footage, amateur-edited clips, and news broadcasts.

In this work, we introduce \dsetname, a benchmark that builds upon MultiVENT and evaluates the ability to write event-centric articles in the style of a Wikipedia lead (overview) section based \emph{only} on video content. Given a request about a real-world event, systems must retrieve a set of relevant videos and then generate an article from the videos' (visual, audio, and OCR) content, providing grounding references (citations) to where the information comes from. \dsetname consists of 57 events and 427 relevant videos (from a corpus of 109K; avg.\ 8 relevant/event), with expert-written reference articles that synthesize information about each event across \emph{all} relevant videos---forcing systems to not only understand high-level information within a \emph{single} video, but also to synthesize information \emph{across} multiple videos on the same topic. Systems able to achieve strong results on \dsetname would be of significant practical use in grounded article generation from multimodal sources, and would enable both the rapid seeding of new Wikipedia articles for actively unfolding events and the enriching of existing articles with 
audiovisual content.

To support the \dsetname task, we propose \methodname (\textbf{C}ollaborative \textbf{A}rticle \textbf{G}eneration), a novel method capable of generating high-level articles. Inspired by relevance feedback \citep{Rocchio1971RelevanceFI} and recent advances in \emph{test-time scaling} \citep{deepseekai2025deepseekr1incentivizingreasoningcapability}, \methodname involves collaborative interaction between (1) a VideoLLM, (2) a text-based reasoning model, and (3) a text-only LLM to extract information from videos and to aggregate the information into an article. The VideoLLM extracts low-level information, such as on-screen text and descriptions of visual entities, while the reasoning model provides relevance feedback on the extracted information, optionally requesting new extractions from the VideoLLM before feeding \emph{relevant} ones to the text-only LLM. The LLM then aggregates the relevant extractions, drawing higher-level inferences about the underlying event, in order to generate the final article. We summarize our contributions as follows:
\begin{enumerate}
    % \vspace{-0.5em}
    \item We introduce \dsetname, a new dataset and task for generating articles from multiple videos. \dsetname is the first benchmark for multi-video article generation, requiring reasoning across audio and visual information, covering 57 events (topics) as depicted in 427 videos that are densely annotated with modality-specific claim grounding annotations, with expert-written reference articles for each event.
    \item We introduce \methodname, a novel method for article generation from multiple videos that is based on relevance feedback and test-time scaling.
    % \vspace{-0.5em}
    \item We present a broad suite of experiments that evaluate both \methodname and popular VideoLLMs on \dsetname across a range of settings, demonstrating \methodname's superiority to other approaches while revealing \dsetname to be a challenging benchmark.
\end{enumerate}

\section{Related Work}
\label{sec:related}
\paragraph{Video Understanding and Summarization}
Video summarization has been studied on small-scale video datasets, such as SumMe \citep[25 videos;][]{gygli2014SumMe} and VideoSum \citep[50 videos;][]{song2015tvsum}---considerably smaller than \dsetname ($\sim400$ videos). Work on \emph{cross-modal} summarization that leverages video largely focuses on producing low-level scene descriptions as the summary, since the associated tasks are chiefly concerned with \emph{aligning} video scenes with caption-like text \citep{he2023bliss, lin2024videoxum, hua2024v2xumllmcrossmodalvideosummarization}. Other work treats video summaries as mere LLM syntheses of frame-level captions \citep{hua2024v2xumllmcrossmodalvideosummarization, zhang2024videoinstructiontuningsynthetic}. In contrast, \dsetname is focused on summaries that provide \emph{high-level} information \emph{supported by} video content.

\citet{ren2025videoragretrievalaugmentedgenerationextreme} recently proposed the task of video retrieval augmented generation (videoRAG) over long videos. While we too explore retrieval in our experiments, our data and task are considerably different: whereas \citeauthor{ren2025videoragretrievalaugmentedgenerationextreme} exclusively use highly polished videos (e.g.\ documentaries, lectures) and only single videos, much of \dsetname consists of raw and amateur-edited footage of events \emph{in the wild} and \emph{in real time} and invovles reasoning across information sources. Further, whereas they are concerned with short-form question answering, we are concerned with long-form article generation.

Similarly, there is much other video understanding work oriented toward tasks other than summarization, such as retrieval of \citep{chen-dolan-2011-collecting, 7780940, Hendricks_2017_ICCV, wang2020vatexlargescalehighqualitymultilingual} question answering about \citep{jang2017tgifqaspatiotemporalreasoningvisual, lei-etal-2018-tvqa, zhou2019activitynetqa}, and recognition of \citep{Zhou_2019_CVPR, sanders-etal-2024-grounding} low-level video features and concepts that span a few seconds or exist only at the frame level.

\paragraph{Article Generation} has largely been studied in text-only settings and as a multi-document summarization task. Early work in this area was performed as part of the DUC and TAC conferences\footnote{DUC: \url{https://duc.nist.gov/}; TAC: \url{https://tac.nist.gov}}, including the DUC 2003-2007 multi-document summarization tasks and the TAC 2010 and 2011 Guided Summarization track. Whereas this early work---and many more recent efforts \citep[][\emph{i.a.}]{hermann2015teaching, nallapati-etal-2016-abstractive, fabbri-etal-2019-multi, huang-etal-2024-embrace}---focused on summarizing \emph{news} articles, generation of Wikipedia-style articles has received increasing attention. To our knowledge, \cite{sauper-barzilay-2009-automatically} were the first to have attempted this, focusing on generation of full Wikipedia articles by filling learned article templates with sentences extracted from source articles. Similar to us, \cite{Liu2018GeneratingWB} focus on Wikipedia lead sections, taking article titles and a collection of source documents as input and performing abstractive summarization using a decoder-only Transformer and introducing the WikiSum dataset as part of their work. \cite{zhu-etal-2021-twag} also focus on leads, but take a topic modeling-inspired approach, assigning topics to source article paragraphs and conditioning generation of each lead sentence on paragraphs associated with either a single predicted topic or a mixture of topics. Like \citet{sauper-barzilay-2009-automatically}, \citet{shao2024assistingwritingwikipedialikearticles} tackle full Wikipedia article generation, using a complex pipeline that entails (1) surveying related Wikipedia articles, (2) generating perspectival questions and answers via simulated dialogues, (3) using these dialogues to construct an outline for the article, and (4) populating the outline from section titles and headings of source articles retrieved during (2). Concurrently, \citet{yang2025wikiautogenmultimodalwikipediastylearticle} explore multimodal article generation, but aimed at incorporating figures into articles rather than synthesizing information.

Finally, in focusing on Wikipedia articles about \emph{events}, we extend a recent line of work on explicitly \emph{event-centric} summarization, in which generations must cover relevant information about a \emph{single} target event \citep{vallurupalli-etal-2022-poque, gantt-etal-2024-event, walden2024crossdocumenteventkeyedsummarization}.

\begin{figure*}[]
    \centering
    \includegraphics[width=\linewidth]{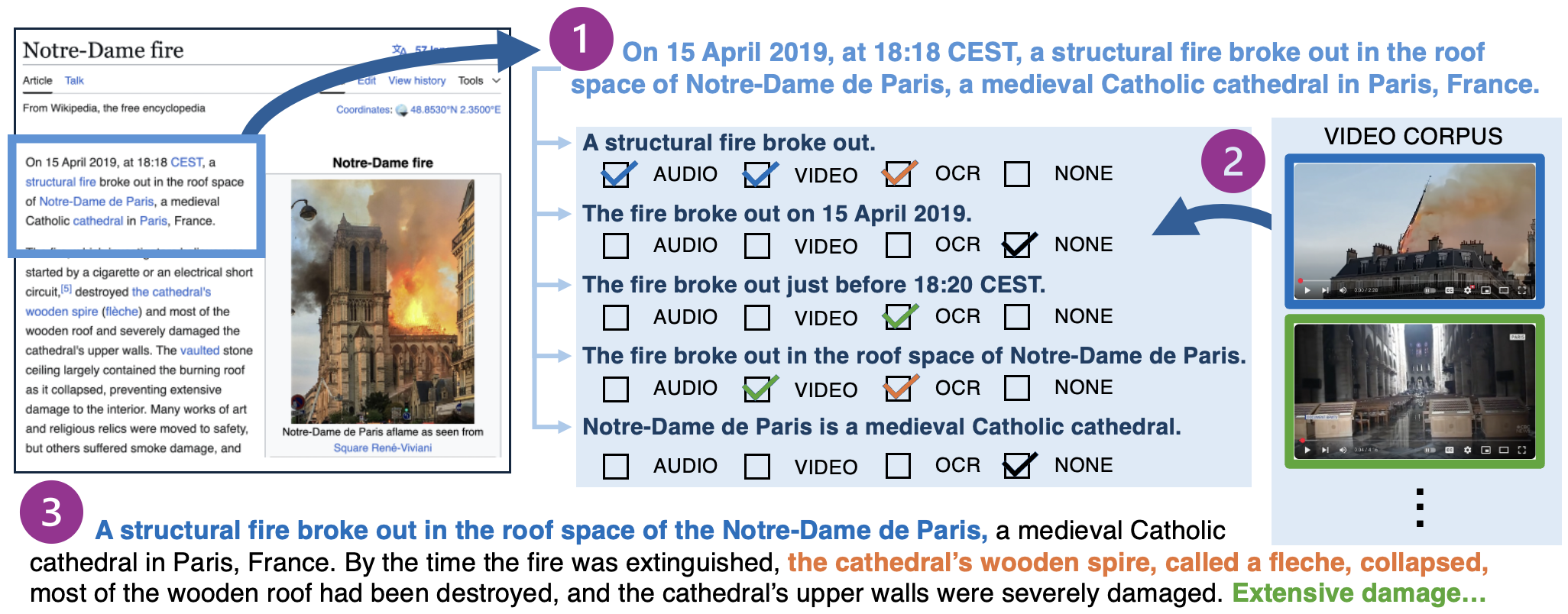}
    \caption{The \dsetname curation process. (1) Sentences in Wikipedia lead sections are decomposed into subclaims. (2) Subclaims are grounded in audio, video, and/or OCR evidence. (3) Leads are rewritten to cover only the \emph{grounded} information.}
    \label{fig:dataset_collection}
    \vspace{-1em}
\end{figure*}

\section{\dsetname Dataset}
\label{sec:data}

\dsetname is built using videos from MultiVENT 1.0 \citep{sanders2023multivent} and MultiVENT 2.0 \citep{kriz2024multivent20massivemultilingual} that are linked to Wikipedia articles obtained from the May 2025 dump provided by the MegaWika2 dataset \citep{barham2025megawika2}. Our data collection process consists of five steps: (1) initial event and article selection, (2) article claim decomposition, (3) claim correction, (4) claim grounding, and (5) article rewriting. \autoref{fig:dataset_collection} illustrates the core components of the annotation process: decomposition, grounding, and rewriting.

\paragraph{Initial Event and Article Selection}
% \textbf{Initial Article Selection:}
We select an initial set of events from MultiVENT subject to two constraints: the event must (1) have English-language videos associated with it and (2) must have a link to an English Wikipedia article about the event.\footnote{As MultiVENT is \emph{multilingual}, not all events it contains satisfy (1) and many also do not satisfy (2).} In total, there are 63 events in MultiVENT satisfying both criteria, supported by 503 associated videos.

For each, we extract the lead section from the linked Wikipedia article to use as the basis for the remainder of our annotation. Lead sections are distinctly suited to our goal of curating \emph{high-level} articles, as they are intended to provide a summary of the most important aspects of the entire page.\footnote{\url{https://en.wikipedia.org/wiki/Wikipedia:Manual_of_Style/Lead_section}}

\paragraph{Claim Decomposition and Correction} 
Next, following recent work on \emph{claim decomposition} \citep[][\emph{i.a.}]{min-etal-2023-factscore, wanner-etal-2024-closer, gunjal-durrett-2024-molecular}, we decompose each sentence of the Wikipedia lead section into a set of contextualized, atomic \emph{subclaims} via few-shot prompting of Qwen 2.5 32B \citep{qwen2025qwen25technicalreport}. 

Expert annotators versed in the claim decomposition literature then manually correct these decompositions to ensure atomicity and faithfulness to the original text.\footnote{\autoref{append:annotation} for details on subclaim decomposition/correction.}

% \begin{wraptable}{r}{5.5cm}
% \centering
% \vspace{-2em}
% \begin{tabular}{lc}
% \toprule
% % \multicolumn{2}{c}{\dsetname}
% \multicolumn{2}{c}{\textbf{\dsetname}} \\
% \midrule
% Video Length (s) & 79.6\\ 
% Article Length (toks) & 118\\
% \midrule
% Videos & 7.65\\
% % \multirow{4}{*}{Claims}
% Audio Subclaims & 25.0\\
% Video Subclaims & 18.3\\
% OCR Subclaims & 28.5 \\
% A/V/O Subclaims & 13.0 \\
% Total Subclaims & 51.1 \\
% \bottomrule
% \end{tabular}
% \caption{Dataset (top) and per-event (bottom) averages.}
% \label{tab:data_stats}
% \end{wraptable} 
\begin{table}[]
    \centering
    \begin{tabular}{lc}
        \toprule
        % \multicolumn{2}{c}{\dsetname}
        \multicolumn{2}{c}{\textbf{\dsetname}} \\
        \midrule
        Video Length (s) & 79.6\\ 
        Article Length (toks) & 118\\
        \midrule
        Videos & 7.65\\
        % \multirow{4}{*}{Claims}
        Audio Subclaims & 25.0\\
        Video Subclaims & 18.3\\
        OCR Subclaims & 28.5 \\
        A/V/O Subclaims & 13.0 \\
        Total Subclaims & 51.1 \\
        \bottomrule
    \end{tabular}
    \caption{Dataset (top) and per-event (bottom) averages.}
    \label{tab:data_stats}
    \vspace{-1em}
\end{table}

% \begin{table}[]
%     \centering
%     \begin{tabular}{c|c}
%     \toprule
%          Video Length (s) & 79.6\\ 
%         Article Length (toks) & 118\\
%         \midrule
%         Videos & 7.5\\
%         % \multirow{4}{*}{Claims}
%         Audio Subclaims & 25.0\\
%         Video Subclaims & 18.3\\
%         OCR Subclaims & 28.5 \\
%         A/V/O Subclaims & 13.0 \\
%         Total Subclaims & 51.1 \\
%     \bottomrule
%     \end{tabular}
%     \caption{Dataset (top) and per-event (bottom) averages.}
%     \label{tab:data_stats}
% \end{table}

% Claims found in audio: 1299
% Claims found in video: 954
% Claims found in both: 674
% Claims not found: 24127
% Total claims: 26782
% Average kept claims per topic: 51.05769230769231
% Average audio claims per topic: 24.98076923076923
% Average video claims per topic: 18.346153846153847
% Average claims found in both: 12.961538461538462
% Max videos: 29
% Min videos: 0
% Average videos: 7.653846153846154

% \vspace{-1em}
\paragraph{Subclaim Grounding}
Given the corrected subclaims for the Wikipedia lead section associated with each event, we next attempt to \emph{ground} the subclaims in relevant videos. This grounding task provides \emph{modality-specific} annotations for each subclaim and for each video associated with the target event, as annotators were asked to indicate whether a subclaim is supported by the video's (non-text) visual content, its OCR content, its audio content, or none of the above. In-domain expert annotators completed these annotations for all 58 events and 503 videos. Our pilot with these annotators obtained a high overall agreement ($\alpha=.767$).

\paragraph{Article Rewriting}
Finally, given the grounded set of subclaims and their corresponding videos, three of the authors rewrote the Wikipedia lead sections such that the resulting articles contained all and only information supported by the video-grounded subclaims. During this stage, six events were found to have too few grounded subclaims to support a rewritten article of any substance, and were subsequently removed from the final dataset.

\paragraph{Final Dataset}
The final \dsetname dataset consists of 57 events (topics) spanning 427 videos annotated with grounded subclaims, where each event is associated with a fully grounded, expert-written article. The events in \dsetname span from 2016-2025, with a specific subset being outside the parametric memory of current (V)LMs. \autoref{tab:data_stats} provides a summary, with more in \autoref{append:stats} \& \ref{append:annotation}.

\section{Article Generation from Multiple Videos}
\label{sec:textgen}

\paragraph{Task} The \dsetname article generation task takes as input (1) a topic event $T$, (2) a query about $T$, and (3) a set of videos $V=\{v_1,\ldots,v_n\}$ deemed relevant to (i.e.\ depicting some facet of) $T$. The output is then a natural language article $A_p$ generated conditional on $T$, $Q$, and $V$. $A_p$ must be fully grounded in $V$, providing citations to videos that support the claims in $A_p$. In this work, we consider two possible sources for $V$: the reference set of videos for $T$ as annotated in MultiVENT 1.0 and 2.0 (the \emph{oracle} setting) and a set of videos obtained from a retrieval model (the \emph{RAG} setting).

\subsection{\underline{C}ollaborative \underline{A}rticle \underline{G}eneration (CAG)}
\paragraph{Overview} Conditional text generation from multiple videos faces several challenges that hinder the efficiency and effectiveness of current methods. First, open-source VideoLLMs are generally trained to produce low-level scene descriptions, making extraction of high-level concepts (necessary for complex event understanding) a challenge, even based on a single video---let alone multiple. Second, running inference over multiple long videos is memory-intensive. For instance, in preliminary experiments with several of the VideoLLMs we consider in even 8 80GB A100s struggled to accommodate a single long video (5+ minutes) at 1 fps, as well as two or more videos at 0.25 fps, additionally supported by the findings of \cite{li2025videochatflashhierarchicalcompressionlongcontext}.

\begin{figure*}[t]
    \centering
    \includegraphics[width=0.9\linewidth]{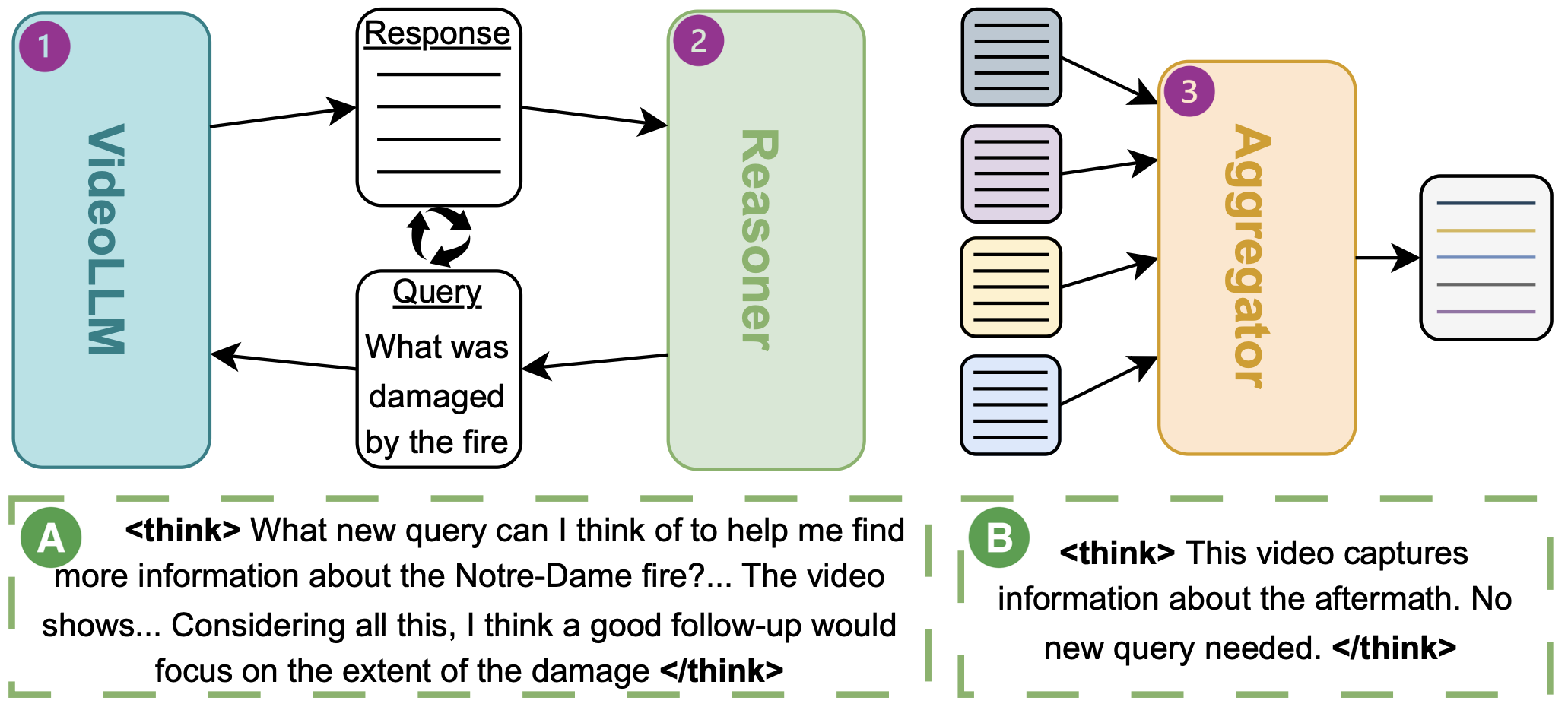}
    \caption{\methodname involves an iterative exchange between (1) a VideoLLM that generates per-video summaries and (2) a reasoning model that evaluates them and produces more event-targeted prompts that are then fed back to the VideoLLM to obtain more comprehensive summaries. Finally, a text-only LLM (3) aggregates these summaries into an full article. Boxes A and B show shortened reasoning chains from the reasoner.
    % \vspace{-5mm}
    }
    \label{fig:method}
\end{figure*}

To help address these limitations, we introduce \textbf{C}ollaborative \textbf{A}rticle \textbf{G}eneration (\methodname; \autoref{fig:method}), a method for article generation from multiple videos that draws on recent developments in \emph{test-time scaling} \citep{deepseekai2025deepseekr1incentivizingreasoningcapability, huang2025visionr1incentivizingreasoningcapability, weller2025rank1testtimecomputereranking, jurayj2025finalanswertesttimescaling} in addition to the classic notion of \emph{relevance feedback} from information retrieval \citep{Rocchio1971RelevanceFI}. \methodname features three core components: a VideoLLM, a reasoning model, and an LLM.

\paragraph{Collaborative Per-Video Summarization}
The first phase of CAG involves a collaborative, iterative exchange between the VideoLLM and the reasoning model. The VideoLLM begins by generating generic summaries of each video based on a simple prompt to ``describe the video in detail.'' The resulting summaries provide salient low-level information, covering scene descriptions and prominent on-screen text.

Next, the reasoning model assists the VideoLLM in producing a \emph{refined} summary for each video that covers higher-level information about the underlying event. Concretely, the reasoning model is given both $Q$ (here, the name of the target event $T$) and the initial generic summary for a particular video as input, and is then asked to either: (1) return the original summary if the reasoning model deems it to be adequate, or else (2) generate a \emph{new} prompt  seeking additional information about $T$ not attested in the input summary. This new prompt is then used to elicit a refined summary from the VideoLLM---an action we dub \textsc{RePrompting}. The reasoning model can thus be understood as providing a form of relevance feedback on the VideoLLM-generated summary with respect to $Q$ (and thus to $T$). This process is iterative because the reasoning model may in principle \textsc{RePrompt} repeatedly---requesting new summaries from the VideoLLM that are more relevant to $Q$ until it is satisfied with the result. In practice, we enforce a maximum \textsc{RePrompting} \emph{iteration budget}---analogous to \emph{test-time compute} budgets for recent reasoning models---that a higher budget tends to yield higher-quality articles.

\paragraph{Article Synthesis}
Once the reasoning model determines that the current query is adequate (or the \textsc{RePrompting} iteration budget is exhausted), the final article is synthesized using a text-only LLM. This model takes as input (1) the original generic summary output by the VideoLLM for each video; (2) all \textsc{RePrompted} VideoLLM queries and their resulting (more event-targeted) summaries; and (3; optionally) an audio transcript of each video. Given these inputs, the model is then instructed to generate the article about $T$.

We note that in virtue of \textsc{RePrompting}---enabling a reasoning model to iteratively craft prompts for the VideoLLM in order to produce summaries more explicitly targeted to an event of interest ($T$)---\methodname goes some way toward mitigating the problem of summaries that are overly focused on low-level descriptions.
Further, in processing one video at a time, \methodname reduces the memory burden of simultaneous multi-video inference.

\section{Experiments}
\label{sec:experiments}

We conduct experiments on \dsetname that (1) benchmark \methodname against baseline VLM approaches; (2) demonstrate the strengths of \methodname over LLM approaches in generalization to unseen events and grounding information in videos; (3)~asses the impact of including raw audio and audio transcripts as input to the text-only LLM; and (4) evaluate the effectiveness of different retrievers in the RAG setting, where videos must first be retrieved. (1-3) are conducted in the \emph{oracle} setting, where $V$ consists of all and only relevant videos.

\paragraph{Models}
For VideoLLMs, we consider LLaVA-Video-72B \citep[``LLaVA-Video'' or ``LV'' in results;][]{zhang2024video}, VAST \citep{chen2024vast}, InternVideo2.5-8B \citep[InternVideo2.5 or IV;][]{wang2025internvideo25empoweringvideomllms}, and QwenVL2.5-72B \citep[Qwen2.5VL or QVL;][]{bai2025qwen25vltechnicalreport}. We use DeepSeek-R1 distilled to Qwen-32B \citep{deepseekai2025deepseekr1incentivizingreasoningcapability} as the reasoning model and Qwen2.5 \cite{qwen2025qwen25technicalreport} as the text-only LLM.

\begin{table*}[t]
    \centering
    \begin{tabular}{cc|cccccc}
    \toprule
        \textbf{Method} & \textbf{VideoLLM} & \textbf{R1} & \textbf{R2} & \textbf{RL} & \textbf{BS} & \textbf{Arg} & \textbf{AS} \\
    \midrule
    \multirow{4}{*}{\concatg}
        % & LLaVA-Video-7B & \xmark & \xmark & - & - & - & 79.35 & - & - \\  
        & LLaVA-Video & 7.34 & 1.60 & 4.78 & 71.99 & 19.31 & 5.08 \\
        % \concatg & VAST & 16.62 & 1.71 & 11.19 & 80.55 & 8.04 & 7.13\\
        % Instruct-V2Xum & \xmark & \xmark & - & - & - & - & - & - \\
        & InternVideo2.5 & 11.85 & 2.32 & 7.90 & 80.78 & 18.33 & 9.53 \\
        & QwenVL2.5 & 11.34 & 3.13 & 7.06 & 81.60 & 23.72 & 8.01 \\
    \midrule
    \multirow{3}{*}{\concatrs}
        & LLaVA-Video & 6.36 & 1.51 & 4.22 & 80.03 & 21.34 & 5.50\\
        % & VAST & \\
        % Instruct-V2Xum & \xmark & \xmark & - & - & - & - & - & - \\
         & InternVideo2.5 & 6.93 & 1.68 & 4.83 & 79.62 & 22.48 & 6.19 \\
        & QwenVL2.5 & 8.38 & 2.71 & 5.49 & 81.94 & 22.89 & 7.17\\
    \midrule
    \multirow{4}{*}{\summg}
        % & LLaVA-Video-7B & 34.87 & 10.72 & 20.18 & 86.44 & 28.09 & 16.34\\
        & LLaVA-Video & 30.02 & 8.68 & 17.59 & 77.59 & 26.21 & 13.51 \\
        % \summg & VAST & 19.55 & 1.45 & 12.40 & 82.21 & 11.23 & 10.87 \\
        & InternVideo2.5 & 32.54 & 8.98 & \underline{19.47} & 85.82 & 25.65 & \textbf{17.58} \\
        % & QwenVL2.5-3B & 32.92 & 10.00 & 19.37 & 85.95 & 27.44 & 16.27 \\
        % & QwenVL2.5-7B & 34.01 & 10.05 & 19.47 & 86.24 & 26.97 & 16.72 \\
        & QwenVL2.5 & 33.58 & \underline{10.15} & 19.15 & \underline{86.18} & \underline{28.97} &  \underline{15.63}\\
    \midrule
    \multirow{3}{*}{\cagr}
        & LLaVA-Video & 33.38 & 10.05 & 19.44 & 84.55 & 28.26 & 15.23 \\
        % & VAST &  - & - & - & - & - & - \\
         & InternVideo2.5 & \underline{33.91} & 9.58 & \textbf{20.07} & 86.13 & 27.01 & 14.23 \\
        & QwenVL2.5 & \textbf{33.96} & \textbf{10.90} & 19.45 & \textbf{86.35} & \textbf{30.77} & 14.29 \\
    % \midrule
    % LLM Only & Qwen2.5 & \\
    
    \bottomrule
    \end{tabular}
    \caption{\dsetname article generation results for \methodname and baselines \emph{without} audio inputs (i.e.\ vision only). \methodname obtains the best results (\textbf{bolded}) across metrics, with performance increasing with \# iterations (2) for most metrics. \vspace{-6mm}}
    \label{tab:main_result}
\end{table*}
\paragraph{Metrics}
We use a suite of different metrics to evaluate the generated articles. We first present ROUGE-\{1,2,LCS\} $\text{F}_1$ \citep[R1, R2, RL;][]{lin-2004-rouge} and BERTScore $\text{F}_1$ \citep[BS;][]{zhang-etal-2019-bertscore} as two widely used metrics for free-form text generation, using the human-written articles as references. As these metrics largely focus on \emph{lexical} similarity, we additionally present AlignScore \citep[AS;][]{zha-etal-2023-alignscore}, a metric for factual consistency based on a learned text-pair alignment function that outputs a scalar value in $[0,1]$ representing the degree of information alignment between the two texts.

Lastly, we also evaluate the extent to which predicted articles recover \emph{specific} pieces of event-relevant information. We map each \dsetname event into the seven-type event ontology defined in the MultiVENT-G dataset \citep{sanders-etal-2024-grounding}, each of which is associated with a set of role-focused questions about events of that type.\footnote{The event types are Sporting Events, Natural Disasters, Elections, Social Events, Demonstrations, Discoveries/Launches, and Political Developments. Event specific scores in \autoref{append:numbers}.} We use an LLM (GPT-4o) to extract answers to these questions from both the reference and predicted articles. Since a question may have multiple answers, we compute a maximum bipartite matching between predicted and reference answers, obtaining an alignment between them that optimizes normalized edit distance between paired answer spans. We then report an answer span $\text{F}_1$ given this alignment, using normalized edit distance in lieu of (overly stringent) exact match. Prior work on event extraction has leveraged similar metrics to evaluate event \emph{argument} $\text{F}_1$ \citep{du-etal-2021-grit, chen-etal-2023-unified, chen-etal-2023-iterative, vashishtha-etal-2024-famus}, so we refer to this as ``Arg.''

\paragraph{Baselines} We consider several baseline article generation methods that ablate different components of \methodname. The first baseline (\concatg) simply concatenates the generic per-video summaries to produce the final article, ablating both the aggregator and reprompting. The second (\concatr) concatenates only the per-video \textsc{RePrompted} summaries, excluding the generic ones, while still ablating the aggregator. The third, (\summg), uses the aggregator but fixes \methodname's iteration budget to 0---relying exclusively on the generic per-video summaries. The comparison between \summg and \cagr (iteration budget of 2) thus offers an illustration of test-time scaling of \methodname via a larger iteration budget.

\subsection{\methodname and Baselines}
\label{subsec:main_experiment} \autoref{tab:main_result} shows \dsetname article generation results comparing \cagr against the baselines described above. We find that simple concatenations of the per-video summaries---whether the initial generic ones (\concatg) or those obtained via reprompting (\concatr, \concatrs)---yield articles of poor quality. Although manual inspection reveals these per-video summaries to contain mostly accurate descriptions of scenes and notable visual entities (e.g.\ the Eiffel tower), we take this as compelling evidence that individual video summaries are inadequate for our task, absent higher-level synthesis.

Results with \summg and \cagr, both of which incorporate the text-only aggregator LLM, offer further evidence for this interpretation, as we observe large gains across all metrics for both of these methods relative to the \textsc{Concat} baselines. For most metrics, \cagr also obtains superior results to \summg, suggesting that supplying the aggregator with the reprompted summaries (in addition to the generic ones) further enhances article quality.

\subsection{Generalization to Unseen Events}
LLMs are trained on vast quantities of internet text, including Wikipedia \cite{barham2023megawikamillionsreportssources, soldaini-etal-2024-dolma, cheng2024dated}. As such, the LLMs underlying each VLM in our experiments can be assumed to have parametric knowledge about WikiVideo events from 2016 to 2024. To illustrate the generalizability of CAG beyond these events, \autoref{tab:2025_qual} reports results on a subset of WikiVideo containing events that occurred \emph{after} the latest knowledge cutoff date of any of these models.\footnote{Qwen2.5 Release: 2024, Earliest 2025 Event: January 15}. Here, we find that \methodname obtains superior results to the Qwen 2.5 LLM, indicating that it exhibits better generalization to genuinely novel events. 

\begin{table}[]
    \centering
    \begin{tabular}{c|ccccc}
    \toprule
         \textbf{Method} & \textbf{R1} & \textbf{BS} & \textbf{Arg} & \textbf{AS} & \textbf{G} \\
    \midrule
        QVL & 15.6 & 82.6 & 25.6 & 15.9 & $-$ \\
        CAG & \textbf{42.0} & \textbf{85.7} & \textbf{32.8} & 14.0 & \textbf{51.2} \\
        \midrule
        LLM & 33.5 & 84.5 & 22.9 & \textbf{18.4} & 19.5 \\ 
    \bottomrule
    \end{tabular}
    \caption{Results on \dsetname-25 (only from 2025). LLM- Qwen 2.5 LLM. G: human annotated grounding.}
    \label{tab:2025_qual}
\end{table}
This experiment also highlights an interesting finding: \emph{VLMs struggle to connect visual signals to parametric knowledge.} This can be see in \autoref{tab:2025_qual}, where the performance of \concatg is similar for all events and the 2025 subset. \autoref{append:outputs} shows summaries for a single video about the 2019 Notre Dame Fire generated by QwenVL2.5 (left) and \methodname (right). While the single video summary correctly identifies the location (Paris), the most salient entity (Notre Dame), and the physical event (fire), the output doesn't make any high-level inference about the event and it fails to connect the visual information to the event in parametric knowledge. We suggest two possible reasons for this: (1) the backbone LLM loses the ability to perform high-level inference because text-video pretraining data is purely focused on low-level descriptions \cite{zhang2024videoinstructiontuningsynthetic}, or (2) the LLM does not have the ability to activate this parametric knowledge from the projected visual signals.

\subsection{Grounding in Videos}
Information presented in the generated articles should be fully grounded in the videos provided as context. To evaluate this, we have human annotators perform the WikiVideo claim grounding annotation task on the model predictions from \autoref{tab:2025_qual} for \methodname and the Qwen2.5 LLM. We score the \emph{G}roundedness of an article as the mean number of claims judged as supported by the video content: $G = \frac{1}{|C|}\sum_{c_i\in C}f(c_i, V)$, where $C$ is the set of claims in the output, $V$ is the set of associated videos, and $f$ is scored as:
\begin{equation}
    f(c, V) = \begin{cases}
        1 & \text{$c$ is supported by some $v_i \in V$} \\
        0 & \text{otherwise}
    \end{cases}
\end{equation}
In \autoref{tab:2025_qual}, we find that a majority of the claims from \methodname are grounded in the article---a stronger performance than the LLM-only generated responses. However, the difference between Rouge and BertScore does not reflect the poor quality of LLM generations or the lack of grounding, highlighting a need for strong evaluation metrics for video-to-text tasks.\footnote{\autoref{append:outputs} for qualitative comparison between \methodname and LLM.}

\subsection{Experiment 5: Including Audio}
\begin{table}
    \centering
    \begin{tabular}{lc|cccc}
    \toprule
         \textbf{Method} & \textbf{VLM} & \textbf{R1} & \textbf{BS} & \textbf{Arg} & \textbf{AS} \\
    \midrule
        % LLaVA-Video-7B & \xmark & \cmark & - & - & - & - & - & - \\  
        \multirow{1}{*}{CAT-0} 
        % & LV & 5.2 & 79.5 & 20.3 & 6.0 \\
        % % Instruct-V2Xum & \xmark & \cmark & - & - & - & - & - & - \\
        % & IV & 11.6 & 80.7 & 18.7 & 8.5 \\
        & QVL & 11.1 & 81.6 & 22.1 & 7.50 \\
        % LLaVA-Video-7B & \cmark & \cmark & - & - & - & - & - & - \\
        \multirow{1}{*}{\cagr}
        % & LV & 29.4 & 77.3 & 25.6 & \underline{13.2} \\
        % % VAST & \cmark & \cmark &  - & - & - & - & - & - \\
        % % Instruct-V2Xum & \cmark & \cmark & - & - & - & - & - & - \\
        %  & IV & \underline{32.8} & \underline{85.8} & 24.8 & 13.1 \\
        & QVL & 32.1 & 85.7 & \underline{26.3} & 12.6 \\
    \midrule
    \rowcolor{gray!15}
        \cagr & QVL & \textbf{34.0} & \textbf{86.4} & \textbf{30.8} & \textbf{14.3} \\
    \bottomrule
    % \rowcolor{gray!25}
    %     Human & & & 40.05 & 13.33 & 20.79 & 86.61 \\
    % \rowcolor{gray!25}
    %     Wikipedia & & & 64.53 & 48.27 & 53.25 & 90.47 \\
    \end{tabular}
    \caption{\dsetname article generation results for \methodname  \emph{with} audio inputs. Bottom row shows \methodname \emph{without} audio (copied from \autoref{tab:main_result}), which performs best.}
    \label{tab:modality_results}
    % \vspace{-0.5em}
\end{table}

Articles in \dsetname have many claims grounded partly or only in videos' audio signal~ (Table~\ref{tab:data_stats}). Here, we consider the impact of adding audio information as additional input to \methodname. In Table~\ref{tab:modality_results} we observe consistently worse results with audio inputs than without. We found that including audio transcripts tended to result in substantially shorter final articles (avg.\ $\sim 164$ tokens vs.\ $\sim 206$ tokens)---suggesting that they may be less thorough in their coverage of the event relative to the references. Identifying ways to more effectively incorporate audio into the \dsetname task is thus an intriguing direction for future work.\footnote{A longer discussion of audio including is in \autoref{append:modality_experiment}}.

\begin{table}
    \centering
    \addtolength{\tabcolsep}{-0.0em}
    \begin{tabular}{cc|ccccccc}
    \toprule
    \textbf{Retriever} & \textbf{VLM} & \textbf{R1} & \textbf{BS} & \textbf{Arg} & \textbf{AS} \\
    \midrule
    % \multirow{5}{*}{\concatg}
    %     % LLaVA-Video-7B & VideoColBERT & \xmark & -  & - & - & - & - & - & - \\
    % & \multirow{2}{*}{V-ColBERT}
    %     % & LV &  \\
    %     % VAST & VideoColBERT & \xmark & - & - & - & - & - & - & -\\
    %     & IV & 12.63 & 2.53 & 8.33 & 80.12 & - & 7.69\\
    %     & & QVL & 11.20 & 3.45 & 6.96 & 81.95 & - & 6.38\\
    % & \multirow{2}{*}{MMMORRF}
    %     % & LV & \\
    %     % VAST & VideoColBERT & \xmark & - & - & - & - & - & - & -\\
    %     & IV & 15.73 & 2.79 & 10.67 & 74.86 & - & 7.43 \\
    %     & & QVL & 16.16 & 4.27 & 9.70 & 75.89 & - & 6.34 \\
    %     \rowcolor{gray!10}
    %     & Oracle & QVL & 11.34 & 3.13 & 7.06 & 81.60 & 23.72 & 8.01 \\
    % \midrule
    % \multirow{5}{*}{\methodname-2}
    % & \multirow{2}{*}{V-ColBERT}
    %     % & LV & \\
    %     & IV & 20.46 & 3.83 & 12.77 & 82.74 & 17.24 &  7.49\\
    %     & & QVL & \underline{24.13} & 4.68 & \underline{14.42} & \underline{83.67} & \underline{20.93} & \underline{10.59} \\
    % & \multirow{2}{*}{\methodana}
        % & LV & \\
    % V-ColBERT & IV & 20.46 & 82.74 & 17.24 &  7.49\\
    % & QVL & \underline{24.13} & \underline{83.67} & \underline{20.93} & \underline{10.59} \\
    \multirow{1}{*}{VC}
    
    % & IV & 20.5 & 82.7 & 17.2 &  7.5 \\
    & QVL & \underline{24.1} & \underline{83.7} & \underline{20.9} & \underline{10.6} \\
    % \midrule
    \multirow{1}{*}{MRF}
        % & IV & 20.5 & 76.6 & 16.7 & 8.8 \\
        & QVL & 23.8 & 77.9 & 20.7 & 9.0 \\
    \midrule
         Oracle & QVL & \textbf{34.0} & \textbf{86.4} & \textbf{30.8} & \textbf{14.3} \\      
    \bottomrule
    \end{tabular}
    \caption{Results with \methodname using different retrievers. The top 5 videos in a ranked list are used for generation. Oracle retrieval results are from Table~\ref{tab:main_result}. VC: Video-ColBERT, MRF: MMMORRF}
    \label{tab:rag_results}
    \vspace{-1em}
\end{table}

\subsection{Experiment 5: Retrieval Augmented Generation}
\label{subsec:rag_experiment} 
In contrast to the previous two experiments, which were run using only relevant videos for each target event (the \emph{oracle} setting), here we consider the \emph{RAG} setting, in which relevant videos must be retrieved. We use the full set of MultiVENT 2.0 \citep{kriz2024multivent20massivemultilingual} videos from the test set as our corpus (109K videos). We perform retrieval from this collection of videos with two retrievers: Video-ColBERT \citep[VC in results;][]{reddy2025videocolbertcontextualizedlateinteraction} and MMMORRF 
\citep[MRF;][]{samuel2025mmmorrfmultimodalmultilingualmodularized}, the state-of-the-art retrieval method on MultiVENT 2.0. We generate articles using the top 5 videos.

\autoref{tab:rag_results} reports article generation results using MMMORRF (nDCG@5: 0.66) and Video-ColBERT (nDCG@5: 0.22). We observe a significant decrease in \methodname performance in moving from oracle retrieval to the RAG setting. This failure falls on the aggregation module of \methodname: the text-only aggregator LLM struggles to include information from \emph{each} video summary, even for irrelevant videos. In such cases, we find that the aggregator usually partitions the lead section into distinct topics instead of writing about the event covered by the (relevant) majority of retrieved videos.

\section{Conclusion}
\label{sec:conclusion}

In this paper we introduce the difficult task of automatically generating Wikipedia-style articles based on multiple videos about real-world events. We collect and release \dsetname, a benchmark of high-quality, expert-written articles grounded in diverse videos, ranging from amateur footage to professional news coverage, which are densely annotated for multimodal support of the articles' claims. Further, since existing systems for video-based summarization tasks are memory-intensive and overly focused on low-level video descriptions, we introduce \textbf{C}ollaborative \textbf{A}rticle \textbf{G}eneration (CAG)---a strong baseline for our task that leverages elements of relevance feedback and test-time scaling to iteratively construct \emph{high-level} event-centric summaries. Our experiments demonstrate the effectiveness of CAG compared to alternative baselines---both in oracle retrieval and RAG settings. While CAG takes a significant step toward addressing the above limitations, future work remains in: (1)~efficient multi-video inference (single inference step), (2)~effectively integrating audio signal into the article generation process, (3)~training VLMs to perform high-level inference, (4)~connecting visual signals to parametric knowledge, and (5)~improving video retrieval for RAG performance.

\section*{Limitations}
\paragraph{Parametric Knowledge}
Models trained on Wikipedia \cite{barham2023megawikamillionsreportssources, cheng2024dated} will be able to artificially inflate scores as quoting from a Wikipedia article leads to high scores across all metrics than human annotators (see \autoref{append:human_analysis}). We suggest reporting results on each subset (\dsetname-24 and \dsetname-25) to help with this limitation. However, when new models come out trained on Wikipedia articles from 2025, the data will again need to be updated.

\paragraph{Computational Costs}
Multi-video inference is an important problem to solve and a limiting factor of our method. To run inference on the 72B version of \methodname, it requires 8 80GB A100s and this is only for \emph{single video} inference. For multi-video this will continue to compound the computational costs and future work should be dedicated to decreasing these costs. 

\section*{Acknowledgments}
This material is based upon work supported by the National Science Foundation Graduate Research Fellowship under Grant No. DGE2139757. Any opinion, findings, and conclusions or recommendations expressed in this material are those of the author(s) and do not necessarily reflect the views of the National Science Foundation.

\bibliography{custom}

\begin{thebibliography}{65}
\providecommand{\natexlab}[1]{#1}

\bibitem[{Anne~Hendricks et~al.(2017)Anne~Hendricks, Wang, Shechtman, Sivic, Darrell, and Russell}]{Hendricks_2017_ICCV}
Lisa Anne~Hendricks, Oliver Wang, Eli Shechtman, Josef Sivic, Trevor Darrell, and Bryan Russell. 2017.
\newblock Localizing moments in video with natural language.
\newblock In \emph{Proceedings of the IEEE International Conference on Computer Vision (ICCV)}.

\bibitem[{Bai et~al.(2025)Bai, Chen, Liu, Wang, Ge, Song, Dang, Wang, Wang, Tang, Zhong, Zhu, Yang, Li, Wan, Wang, Ding, Fu, Xu, Ye, Zhang, Xie, Cheng, Zhang, Yang, Xu, and Lin}]{bai2025qwen25vltechnicalreport}
Shuai Bai, Keqin Chen, Xuejing Liu, Jialin Wang, Wenbin Ge, Sibo Song, Kai Dang, Peng Wang, Shijie Wang, Jun Tang, Humen Zhong, Yuanzhi Zhu, Mingkun Yang, Zhaohai Li, Jianqiang Wan, Pengfei Wang, Wei Ding, Zheren Fu, Yiheng Xu, and 8 others. 2025.
\newblock \href {https://arxiv.org/abs/2502.13923} {Qwen2.5-vl technical report}.
\newblock \emph{Preprint}, arXiv:2502.13923.

\bibitem[{Barham et~al.(2025)Barham, May, and Durme}]{barham2025megawika2}
Samuel Barham, Chandler May, and Benjamin~Van Durme. 2025.
\newblock \href {https://arxiv.org/abs/2508.03828} {Megawika 2: A more comprehensive multilingual collection of articles and their sources}.
\newblock \emph{Preprint}, arXiv:2508.03828.

\bibitem[{Barham et~al.(2023)Barham, Weller, Yuan, Murray, Yarmohammadi, Jiang, Vashishtha, Martin, Liu, White, Boyd-Graber, and Durme}]{barham2023megawikamillionsreportssources}
Samuel Barham, Orion Weller, Michelle Yuan, Kenton Murray, Mahsa Yarmohammadi, Zhengping Jiang, Siddharth Vashishtha, Alexander Martin, Anqi Liu, Aaron~Steven White, Jordan Boyd-Graber, and Benjamin~Van Durme. 2023.
\newblock \href {https://arxiv.org/abs/2307.07049} {Megawika: Millions of reports and their sources across 50 diverse languages}.
\newblock \emph{Preprint}, arXiv:2307.07049.

\bibitem[{Chen and Dolan(2011)}]{chen-dolan-2011-collecting}
David Chen and William Dolan. 2011.
\newblock \href {https://aclanthology.org/P11-1020/} {Collecting highly parallel data for paraphrase evaluation}.
\newblock In \emph{Proceedings of the 49th Annual Meeting of the Association for Computational Linguistics: Human Language Technologies}, pages 190--200, Portland, Oregon, USA. Association for Computational Linguistics.

\bibitem[{Chen et~al.(2024)Chen, Li, Wang, Zhao, Sun, Zhu, and Liu}]{chen2024vast}
Sihan Chen, Handong Li, Qunbo Wang, Zijia Zhao, Mingzhen Sun, Xinxin Zhu, and Jing Liu. 2024.
\newblock Vast: A vision-audio-subtitle-text omni-modality foundation model and dataset.
\newblock \emph{Advances in Neural Information Processing Systems}, 36.

\bibitem[{Chen et~al.(2023{\natexlab{a}})Chen, Gantt, Chen, White, and Van~Durme}]{chen-etal-2023-unified}
Yunmo Chen, William Gantt, Tongfei Chen, Aaron White, and Benjamin Van~Durme. 2023{\natexlab{a}}.
\newblock \href {https://doi.org/10.18653/v1/2023.emnlp-main.795} {A unified view of evaluation metrics for structured prediction}.
\newblock In \emph{Proceedings of the 2023 Conference on Empirical Methods in Natural Language Processing}, pages 12868--12882, Singapore. Association for Computational Linguistics.

\bibitem[{Chen et~al.(2023{\natexlab{b}})Chen, Gantt, Gu, Chen, White, and Van~Durme}]{chen-etal-2023-iterative}
Yunmo Chen, William Gantt, Weiwei Gu, Tongfei Chen, Aaron White, and Benjamin Van~Durme. 2023{\natexlab{b}}.
\newblock \href {https://doi.org/10.18653/v1/2023.eacl-main.136} {Iterative document-level information extraction via imitation learning}.
\newblock In \emph{Proceedings of the 17th Conference of the European Chapter of the Association for Computational Linguistics}, pages 1858--1874, Dubrovnik, Croatia. Association for Computational Linguistics.

\bibitem[{Cheng et~al.(2024)Cheng, Marone, Weller, Lawrie, Khashabi, and Durme}]{cheng2024dated}
Jeffrey Cheng, Marc Marone, Orion Weller, Dawn Lawrie, Daniel Khashabi, and Benjamin~Van Durme. 2024.
\newblock \href {https://openreview.net/forum?id=wS7PxDjy6m} {Dated data: Tracing knowledge cutoffs in large language models}.
\newblock In \emph{First Conference on Language Modeling}.

\bibitem[{DeepSeek-AI et~al.(2025)DeepSeek-AI, Guo, Yang, Zhang, Song, Zhang, Xu, Zhu, Ma, Wang, Bi, Zhang, Yu, Wu, Wu, Gou, Shao, Li, Gao, Liu, Xue, Wang, Wu, Feng, Lu, Zhao, Deng, Zhang, Ruan, Dai, Chen, Ji, Li, Lin, Dai, Luo, Hao, Chen, Li, Zhang, Bao, Xu, Wang, Ding, Xin, Gao, Qu, Li, Guo, Li, Wang, Chen, Yuan, Qiu, Li, Cai, Ni, Liang, Chen, Dong, Hu, Gao, Guan, Huang, Yu, Wang, Zhang, Zhao, Wang, Zhang, Xu, Xia, Zhang, Zhang, Tang, Li, Wang, Li, Tian, Huang, Zhang, Wang, Chen, Du, Ge, Zhang, Pan, Wang, Chen, Jin, Chen, Lu, Zhou, Chen, Ye, Wang, Yu, Zhou, Pan, Li, Zhou, Wu, Ye, Yun, Pei, Sun, Wang, Zeng, Zhao, Liu, Liang, Gao, Yu, Zhang, Xiao, An, Liu, Wang, Chen, Nie, Cheng, Liu, Xie, Liu, Yang, Li, Su, Lin, Li, Jin, Shen, Chen, Sun, Wang, Song, Zhou, Wang, Shan, Li, Wang, Wei, Zhang, Xu, Li, Zhao, Sun, Wang, Yu, Zhang, Shi, Xiong, He, Piao, Wang, Tan, Ma, Liu, Guo, Ou, Wang, Gong, Zou, He, Xiong, Luo, You, Liu, Zhou, Zhu, Xu, Huang, Li, Zheng, Zhu, Ma, Tang, Zha, Yan, Ren, Ren, Sha, Fu, Xu, Xie, Zhang,
  Hao, Ma, Yan, Wu, Gu, Zhu, Liu, Li, Xie, Song, Pan, Huang, Xu, Zhang, and Zhang}]{deepseekai2025deepseekr1incentivizingreasoningcapability}
DeepSeek-AI, Daya Guo, Dejian Yang, Haowei Zhang, Junxiao Song, Ruoyu Zhang, Runxin Xu, Qihao Zhu, Shirong Ma, Peiyi Wang, Xiao Bi, Xiaokang Zhang, Xingkai Yu, Yu~Wu, Z.~F. Wu, Zhibin Gou, Zhihong Shao, Zhuoshu Li, Ziyi Gao, and 181 others. 2025.
\newblock \href {https://arxiv.org/abs/2501.12948} {Deepseek-r1: Incentivizing reasoning capability in llms via reinforcement learning}.
\newblock \emph{Preprint}, arXiv:2501.12948.

\bibitem[{Du et~al.(2021)Du, Rush, and Cardie}]{du-etal-2021-grit}
Xinya Du, Alexander Rush, and Claire Cardie. 2021.
\newblock \href {https://doi.org/10.18653/v1/2021.eacl-main.52} {{GRIT}: Generative role-filler transformers for document-level event entity extraction}.
\newblock In \emph{Proceedings of the 16th Conference of the European Chapter of the Association for Computational Linguistics: Main Volume}, pages 634--644, Online. Association for Computational Linguistics.

\bibitem[{Fabbri et~al.(2019)Fabbri, Li, She, Li, and Radev}]{fabbri-etal-2019-multi}
Alexander Fabbri, Irene Li, Tianwei She, Suyi Li, and Dragomir Radev. 2019.
\newblock \href {https://doi.org/10.18653/v1/P19-1102} {Multi-news: A large-scale multi-document summarization dataset and abstractive hierarchical model}.
\newblock In \emph{Proceedings of the 57th Annual Meeting of the Association for Computational Linguistics}, pages 1074--1084, Florence, Italy. Association for Computational Linguistics.

\bibitem[{Gantt et~al.(2024)Gantt, Martin, Kuchmiichuk, and White}]{gantt-etal-2024-event}
William Gantt, Alexander Martin, Pavlo Kuchmiichuk, and Aaron~Steven White. 2024.
\newblock \href {https://doi.org/10.18653/v1/2024.findings-emnlp.431} {Event-keyed summarization}.
\newblock In \emph{Findings of the Association for Computational Linguistics: EMNLP 2024}, pages 7333--7345, Miami, Florida, USA. Association for Computational Linguistics.

\bibitem[{Gunjal and Durrett(2024)}]{gunjal-durrett-2024-molecular}
Anisha Gunjal and Greg Durrett. 2024.
\newblock \href {https://doi.org/10.18653/v1/2024.findings-emnlp.215} {Molecular facts: Desiderata for decontextualization in {LLM} fact verification}.
\newblock In \emph{Findings of the Association for Computational Linguistics: EMNLP 2024}, pages 3751--3768, Miami, Florida, USA. Association for Computational Linguistics.

\bibitem[{Gygli et~al.(2014)Gygli, Grabner, Riemenschneider, and Van~Gool}]{gygli2014SumMe}
Michael Gygli, Helmut Grabner, Hayko Riemenschneider, and Luc Van~Gool. 2014.
\newblock Creating summaries from user videos.
\newblock In \emph{Computer Vision -- ECCV 2014}, pages 505--520, Cham. Springer International Publishing.

\bibitem[{He et~al.(2023)He, Wang, Qiu, Bui, Shrivastava, and Wang}]{he2023bliss}
Bo~He, Jun Wang, Jielin Qiu, Trung Bui, Abhinav Shrivastava, and Zhaowen Wang. 2023.
\newblock \href {https://doi.org/10.1109/CVPR52729.2023.01428} {{ Align and Attend: Multimodal Summarization with Dual Contrastive Losses }}.
\newblock In \emph{2023 IEEE/CVF Conference on Computer Vision and Pattern Recognition (CVPR)}, pages 14867--14878, Los Alamitos, CA, USA. IEEE Computer Society.

\bibitem[{Hermann et~al.(2015)Hermann, Kocisky, Grefenstette, Espeholt, Kay, Suleyman, and Blunsom}]{hermann2015teaching}
Karl~Moritz Hermann, Tomas Kocisky, Edward Grefenstette, Lasse Espeholt, Will Kay, Mustafa Suleyman, and Phil Blunsom. 2015.
\newblock Teaching machines to read and comprehend.
\newblock \emph{Advances in neural information processing systems}, 28.

\bibitem[{Hu et~al.(2025)Hu, Long, and Wang}]{hu2025decompositiondilemmasdoesclaim}
Qisheng Hu, Quanyu Long, and Wenya Wang. 2025.
\newblock \href {https://arxiv.org/abs/2411.02400} {Decomposition dilemmas: Does claim decomposition boost or burden fact-checking performance?}
\newblock \emph{Preprint}, arXiv:2411.02400.

\bibitem[{Hua et~al.(2024)Hua, Tang, Xu, and Luo}]{hua2024v2xumllmcrossmodalvideosummarization}
Hang Hua, Yunlong Tang, Chenliang Xu, and Jiebo Luo. 2024.
\newblock \href {https://arxiv.org/abs/2404.12353} {V2xum-llm: Cross-modal video summarization with temporal prompt instruction tuning}.
\newblock \emph{Preprint}, arXiv:2404.12353.

\bibitem[{Huang et~al.(2024)Huang, Laban, Fabbri, Choubey, Joty, Xiong, and Wu}]{huang-etal-2024-embrace}
Kung-Hsiang Huang, Philippe Laban, Alexander Fabbri, Prafulla~Kumar Choubey, Shafiq Joty, Caiming Xiong, and Chien-Sheng Wu. 2024.
\newblock \href {https://doi.org/10.18653/v1/2024.naacl-long.32} {Embrace divergence for richer insights: A multi-document summarization benchmark and a case study on summarizing diverse information from news articles}.
\newblock In \emph{Proceedings of the 2024 Conference of the North American Chapter of the Association for Computational Linguistics: Human Language Technologies (Volume 1: Long Papers)}, pages 570--593, Mexico City, Mexico. Association for Computational Linguistics.

\bibitem[{Huang et~al.(2025)Huang, Jia, Zhai, Cao, Ye, Zhao, Xu, Hu, and Lin}]{huang2025visionr1incentivizingreasoningcapability}
Wenxuan Huang, Bohan Jia, Zijie Zhai, Shaosheng Cao, Zheyu Ye, Fei Zhao, Zhe Xu, Yao Hu, and Shaohui Lin. 2025.
\newblock \href {https://arxiv.org/abs/2503.06749} {Vision-r1: Incentivizing reasoning capability in multimodal large language models}.
\newblock \emph{Preprint}, arXiv:2503.06749.

\bibitem[{Jang et~al.(2017)Jang, Song, Yu, Kim, and Kim}]{jang2017tgifqaspatiotemporalreasoningvisual}
Yunseok Jang, Yale Song, Youngjae Yu, Youngjin Kim, and Gunhee Kim. 2017.
\newblock \href {https://arxiv.org/abs/1704.04497} {Tgif-qa: Toward spatio-temporal reasoning in visual question answering}.
\newblock \emph{Preprint}, arXiv:1704.04497.

\bibitem[{Jurayj et~al.(2025)Jurayj, Cheng, and Durme}]{jurayj2025finalanswertesttimescaling}
William Jurayj, Jeffrey Cheng, and Benjamin~Van Durme. 2025.
\newblock \href {https://arxiv.org/abs/2502.13962} {Is that your final answer? test-time scaling improves selective question answering}.
\newblock \emph{Preprint}, arXiv:2502.13962.

\bibitem[{Krishna et~al.(2017)Krishna, Hata, Ren, Fei-Fei, and Carlos~Niebles}]{Krishna_2017_ICCV}
Ranjay Krishna, Kenji Hata, Frederic Ren, Li~Fei-Fei, and Juan Carlos~Niebles. 2017.
\newblock Dense-captioning events in videos.
\newblock In \emph{Proceedings of the IEEE International Conference on Computer Vision (ICCV)}.

\bibitem[{Kriz et~al.(2024)Kriz, Sanders, Etter, Murray, Carpenter, Ochten, Recknor, Guallar-Blasco, Martin, Colaianni, King, Yang, and Durme}]{kriz2024multivent20massivemultilingual}
Reno Kriz, Kate Sanders, David Etter, Kenton Murray, Cameron Carpenter, Kelly~Van Ochten, Hannah Recknor, Jimena Guallar-Blasco, Alexander Martin, Ronald Colaianni, Nolan King, Eugene Yang, and Benjamin~Van Durme. 2024.
\newblock \href {https://arxiv.org/abs/2410.11619} {Multivent 2.0: A massive multilingual benchmark for event-centric video retrieval}.
\newblock \emph{Preprint}, arXiv:2410.11619.

\bibitem[{Lawrie et~al.(2024)Lawrie, MacAvaney, Mayfield, McNamee, Oard, and åand Eugene~Yang}]{lawrie2024overviewtrec2023neuclir}
Dawn Lawrie, Sean MacAvaney, James Mayfield, Paul McNamee, Douglas~W. Oard, and Luca~Soldaini åand Eugene~Yang. 2024.
\newblock \href {https://arxiv.org/abs/2404.08071} {Overview of the trec 2023 neuclir track}.
\newblock \emph{Preprint}, arXiv:2404.08071.

\bibitem[{Lei et~al.(2018)Lei, Yu, Bansal, and Berg}]{lei-etal-2018-tvqa}
Jie Lei, Licheng Yu, Mohit Bansal, and Tamara Berg. 2018.
\newblock \href {https://doi.org/10.18653/v1/D18-1167} {{TVQA}: Localized, compositional video question answering}.
\newblock In \emph{Proceedings of the 2018 Conference on Empirical Methods in Natural Language Processing}, pages 1369--1379, Brussels, Belgium. Association for Computational Linguistics.

\bibitem[{Li et~al.(2022)Li, Wei, Tian, Xu, Wen, and Hu}]{Li_2022_CVPR}
Guangyao Li, Yake Wei, Yapeng Tian, Chenliang Xu, Ji-Rong Wen, and Di~Hu. 2022.
\newblock Learning to answer questions in dynamic audio-visual scenarios.
\newblock In \emph{Proceedings of the IEEE/CVF Conference on Computer Vision and Pattern Recognition (CVPR)}, pages 19108--19118.

\bibitem[{Li et~al.(2025)Li, Wang, Yu, Zeng, Zhu, Huang, Gao, Li, He, Wang, Qiao, Wang, and Wang}]{li2025videochatflashhierarchicalcompressionlongcontext}
Xinhao Li, Yi~Wang, Jiashuo Yu, Xiangyu Zeng, Yuhan Zhu, Haian Huang, Jianfei Gao, Kunchang Li, Yinan He, Chenting Wang, Yu~Qiao, Yali Wang, and Limin Wang. 2025.
\newblock \href {https://arxiv.org/abs/2501.00574} {Videochat-flash: Hierarchical compression for long-context video modeling}.
\newblock \emph{Preprint}, arXiv:2501.00574.

\bibitem[{Lin(2004)}]{lin-2004-rouge}
Chin-Yew Lin. 2004.
\newblock \href {https://aclanthology.org/W04-1013/} {{ROUGE}: A package for automatic evaluation of summaries}.
\newblock In \emph{Text Summarization Branches Out}, pages 74--81, Barcelona, Spain. Association for Computational Linguistics.

\bibitem[{Lin et~al.(2024)Lin, Hua, Chen, Li, Hsiao, Ho, and Luo}]{lin2024videoxum}
Jingyang Lin, Hang Hua, Ming Chen, Yikang Li, Jenhao Hsiao, Chiuman Ho, and Jiebo Luo. 2024.
\newblock \href {https://doi.org/10.1109/TMM.2023.3335875} {Videoxum: Cross-modal visual and textural summarization of videos}.
\newblock \emph{IEEE Transactions on Multimedia}, 26:5548--5560.

\bibitem[{Liu et~al.(2018{\natexlab{a}})Liu, Saleh, Pot, Goodrich, Sepassi, Kaiser, and Shazeer}]{liu2018generating}
Peter~J. Liu, Mohammad Saleh, Etienne Pot, Ben Goodrich, Ryan Sepassi, Lukasz Kaiser, and Noam Shazeer. 2018{\natexlab{a}}.
\newblock \href {https://openreview.net/forum?id=Hyg0vbWC-} {Generating wikipedia by summarizing long sequences}.
\newblock In \emph{International Conference on Learning Representations}.

\bibitem[{Liu et~al.(2018{\natexlab{b}})Liu, Saleh, Pot, Goodrich, Sepassi, Kaiser, and Shazeer}]{Liu2018GeneratingWB}
Peter~J. Liu, Mohammad Saleh, Etienne Pot, Ben Goodrich, Ryan Sepassi, Lukasz Kaiser, and Noam~M. Shazeer. 2018{\natexlab{b}}.
\newblock \href {https://api.semanticscholar.org/CorpusID:3608234} {Generating wikipedia by summarizing long sequences}.
\newblock \emph{ArXiv}, abs/1801.10198.

\bibitem[{Min et~al.(2023)Min, Krishna, Lyu, Lewis, Yih, Koh, Iyyer, Zettlemoyer, and Hajishirzi}]{min-etal-2023-factscore}
Sewon Min, Kalpesh Krishna, Xinxi Lyu, Mike Lewis, Wen-tau Yih, Pang Koh, Mohit Iyyer, Luke Zettlemoyer, and Hannaneh Hajishirzi. 2023.
\newblock \href {https://doi.org/10.18653/v1/2023.emnlp-main.741} {{FA}ct{S}core: Fine-grained atomic evaluation of factual precision in long form text generation}.
\newblock In \emph{Proceedings of the 2023 Conference on Empirical Methods in Natural Language Processing}, pages 12076--12100, Singapore. Association for Computational Linguistics.

\bibitem[{Nallapati et~al.(2016)Nallapati, Zhou, dos Santos, Gu{\ensuremath{\dot{}}}l{\c{c}}ehre, and Xiang}]{nallapati-etal-2016-abstractive}
Ramesh Nallapati, Bowen Zhou, Cicero dos Santos, {\c{C}}a{\u{g}}lar Gu{\ensuremath{\dot{}}}l{\c{c}}ehre, and Bing Xiang. 2016.
\newblock \href {https://doi.org/10.18653/v1/K16-1028} {Abstractive text summarization using sequence-to-sequence {RNN}s and beyond}.
\newblock In \emph{Proceedings of the 20th {SIGNLL} Conference on Computational Natural Language Learning}, pages 280--290, Berlin, Germany. Association for Computational Linguistics.

\bibitem[{Qwen et~al.(2025)Qwen, :, Yang, Yang, Zhang, Hui, Zheng, Yu, Li, Liu, Huang, Wei, Lin, Yang, Tu, Zhang, Yang, Yang, Zhou, Lin, Dang, Lu, Bao, Yang, Yu, Li, Xue, Zhang, Zhu, Men, Lin, Li, Tang, Xia, Ren, Ren, Fan, Su, Zhang, Wan, Liu, Cui, Zhang, and Qiu}]{qwen2025qwen25technicalreport}
Qwen, :, An~Yang, Baosong Yang, Beichen Zhang, Binyuan Hui, Bo~Zheng, Bowen Yu, Chengyuan Li, Dayiheng Liu, Fei Huang, Haoran Wei, Huan Lin, Jian Yang, Jianhong Tu, Jianwei Zhang, Jianxin Yang, Jiaxi Yang, Jingren Zhou, and 25 others. 2025.
\newblock \href {https://arxiv.org/abs/2412.15115} {Qwen2.5 technical report}.
\newblock \emph{Preprint}, arXiv:2412.15115.

\bibitem[{Radford et~al.(2022)Radford, Kim, Xu, Brockman, McLeavey, and Sutskever}]{radford2022whisper}
Alec Radford, Jong~Wook Kim, Tao Xu, Greg Brockman, Christine McLeavey, and Ilya Sutskever. 2022.
\newblock \href {https://doi.org/10.48550/ARXIV.2212.04356} {Robust speech recognition via large-scale weak supervision}.
\newblock \emph{arXiv preprint}.

\bibitem[{Reddy et~al.(2025)Reddy, Martin, Yang, Yates, Sanders, Murray, Kriz, de~Melo, Durme, and Chellappa}]{reddy2025videocolbertcontextualizedlateinteraction}
Arun Reddy, Alexander Martin, Eugene Yang, Andrew Yates, Kate Sanders, Kenton Murray, Reno Kriz, Celso~M. de~Melo, Benjamin~Van Durme, and Rama Chellappa. 2025.
\newblock \href {https://arxiv.org/abs/2503.19009} {Video-colbert: Contextualized late interaction for text-to-video retrieval}.
\newblock \emph{Preprint}, arXiv:2503.19009.

\bibitem[{Ren et~al.(2025)Ren, Xu, Xia, Wang, Yin, and Huang}]{ren2025videoragretrievalaugmentedgenerationextreme}
Xubin Ren, Lingrui Xu, Long Xia, Shuaiqiang Wang, Dawei Yin, and Chao Huang. 2025.
\newblock \href {https://arxiv.org/abs/2502.01549} {Videorag: Retrieval-augmented generation with extreme long-context videos}.
\newblock \emph{Preprint}, arXiv:2502.01549.

\bibitem[{Rocchio(1971)}]{Rocchio1971RelevanceFI}
J.~J. Rocchio. 1971.
\newblock \href {https://api.semanticscholar.org/CorpusID:61859400} {Relevance feedback in information retrieval}.

\bibitem[{Samuel et~al.(2025)Samuel, DeGenaro, Guallar-Blasco, Sanders, Eisape, Reddy, Martin, Yates, Yang, Carpenter, Etter, Kayi, Wiesner, Murray, and Kriz}]{samuel2025mmmorrfmultimodalmultilingualmodularized}
Saron Samuel, Dan DeGenaro, Jimena Guallar-Blasco, Kate Sanders, Oluwaseun Eisape, Arun Reddy, Alexander Martin, Andrew Yates, Eugene Yang, Cameron Carpenter, David Etter, Efsun Kayi, Matthew Wiesner, Kenton Murray, and Reno Kriz. 2025.
\newblock \href {https://arxiv.org/abs/2503.20698} {Mmmorrf: Multimodal multilingual modularized reciprocal rank fusion}.
\newblock \emph{Preprint}, arXiv:2503.20698.

\bibitem[{Sanders et~al.(2023)Sanders, Etter, Kriz, and Durme}]{sanders2023multivent}
Kate Sanders, David Etter, Reno Kriz, and Benjamin~Van Durme. 2023.
\newblock \href {https://openreview.net/forum?id=2CJUQe6IoR} {Multi{VENT}: Multilingual videos of events and aligned natural text}.
\newblock In \emph{Thirty-seventh Conference on Neural Information Processing Systems Datasets and Benchmarks Track}.

\bibitem[{Sanders et~al.(2024)Sanders, Kriz, Etter, Recknor, Martin, Carpenter, Lin, and Van~Durme}]{sanders-etal-2024-grounding}
Kate Sanders, Reno Kriz, David Etter, Hannah Recknor, Alexander Martin, Cameron Carpenter, Jingyang Lin, and Benjamin Van~Durme. 2024.
\newblock \href {https://doi.org/10.18653/v1/2024.findings-emnlp.934} {Grounding partially-defined events in multimodal data}.
\newblock In \emph{Findings of the Association for Computational Linguistics: EMNLP 2024}, pages 15905--15927, Miami, Florida, USA. Association for Computational Linguistics.

\bibitem[{Sauper and Barzilay(2009)}]{sauper-barzilay-2009-automatically}
Christina Sauper and Regina Barzilay. 2009.
\newblock \href {https://aclanthology.org/P09-1024/} {Automatically generating {W}ikipedia articles: A structure-aware approach}.
\newblock In \emph{Proceedings of the Joint Conference of the 47th Annual Meeting of the {ACL} and the 4th International Joint Conference on Natural Language Processing of the {AFNLP}}, pages 208--216, Suntec, Singapore. Association for Computational Linguistics.

\bibitem[{Shao et~al.(2024)Shao, Jiang, Kanell, Xu, Khattab, and Lam}]{shao2024assistingwritingwikipedialikearticles}
Yijia Shao, Yucheng Jiang, Theodore~A. Kanell, Peter Xu, Omar Khattab, and Monica~S. Lam. 2024.
\newblock \href {https://arxiv.org/abs/2402.14207} {Assisting in writing wikipedia-like articles from scratch with large language models}.
\newblock \emph{Preprint}, arXiv:2402.14207.

\bibitem[{Soldaini et~al.(2024)Soldaini, Kinney, Bhagia, Schwenk, Atkinson, Authur, Bogin, Chandu, Dumas, Elazar, Hofmann, Jha, Kumar, Lucy, Lyu, Lambert, Magnusson, Morrison, Muennighoff, Naik, Nam, Peters, Ravichander, Richardson, Shen, Strubell, Subramani, Tafjord, Walsh, Zettlemoyer, Smith, Hajishirzi, Beltagy, Groeneveld, Dodge, and Lo}]{soldaini-etal-2024-dolma}
Luca Soldaini, Rodney Kinney, Akshita Bhagia, Dustin Schwenk, David Atkinson, Russell Authur, Ben Bogin, Khyathi Chandu, Jennifer Dumas, Yanai Elazar, Valentin Hofmann, Ananya Jha, Sachin Kumar, Li~Lucy, Xinxi Lyu, Nathan Lambert, Ian Magnusson, Jacob Morrison, Niklas Muennighoff, and 17 others. 2024.
\newblock \href {https://doi.org/10.18653/v1/2024.acl-long.840} {Dolma: an open corpus of three trillion tokens for language model pretraining research}.
\newblock In \emph{Proceedings of the 62nd Annual Meeting of the Association for Computational Linguistics (Volume 1: Long Papers)}, pages 15725--15788, Bangkok, Thailand. Association for Computational Linguistics.

\bibitem[{Song et~al.(2015)Song, Vallmitjana, Stent, and Jaimes}]{song2015tvsum}
Yale Song, Jordi Vallmitjana, Amanda Stent, and Alejandro Jaimes. 2015.
\newblock \href {https://doi.org/10.1109/CVPR.2015.7299154} {Tvsum: Summarizing web videos using titles}.
\newblock In \emph{2015 IEEE Conference on Computer Vision and Pattern Recognition (CVPR)}, pages 5179--5187.

\bibitem[{Srikanth and Rudinger(2025)}]{srikanth2025nlimicroscopeatomichypothesis}
Neha Srikanth and Rachel Rudinger. 2025.
\newblock \href {https://arxiv.org/abs/2502.08080} {Nli under the microscope: What atomic hypothesis decomposition reveals}.
\newblock \emph{Preprint}, arXiv:2502.08080.

\bibitem[{Vallurupalli et~al.(2022)Vallurupalli, Ghosh, Erk, Balasubramanian, and Ferraro}]{vallurupalli-etal-2022-poque}
Sai Vallurupalli, Sayontan Ghosh, Katrin Erk, Niranjan Balasubramanian, and Francis Ferraro. 2022.
\newblock \href {https://doi.org/10.18653/v1/2022.emnlp-main.594} {{POQ}ue: Asking participant-specific outcome questions for a deeper understanding of complex events}.
\newblock In \emph{Proceedings of the 2022 Conference on Empirical Methods in Natural Language Processing}, pages 8674--8697, Abu Dhabi, United Arab Emirates. Association for Computational Linguistics.

\bibitem[{Vashishtha et~al.(2024)Vashishtha, Martin, Gantt, Van~Durme, and White}]{vashishtha-etal-2024-famus}
Siddharth Vashishtha, Alexander Martin, William Gantt, Benjamin Van~Durme, and Aaron White. 2024.
\newblock \href {https://doi.org/10.18653/v1/2024.naacl-long.457} {{FAM}u{S}: Frames across multiple sources}.
\newblock In \emph{Proceedings of the 2024 Conference of the North American Chapter of the Association for Computational Linguistics: Human Language Technologies (Volume 1: Long Papers)}, pages 8250--8273, Mexico City, Mexico. Association for Computational Linguistics.

\bibitem[{Walden et~al.(2024)Walden, Kuchmiichuk, Martin, Jin, Cao, Sun, Allen, and White}]{walden2024crossdocumenteventkeyedsummarization}
William Walden, Pavlo Kuchmiichuk, Alexander Martin, Chihsheng Jin, Angela Cao, Claire Sun, Curisia Allen, and Aaron~Steven White. 2024.
\newblock \href {https://arxiv.org/abs/2410.14795} {Cross-document event-keyed summarization}.
\newblock \emph{Preprint}, arXiv:2410.14795.

\bibitem[{Wang et~al.(2020)Wang, Wu, Chen, Li, Wang, and Wang}]{wang2020vatexlargescalehighqualitymultilingual}
Xin Wang, Jiawei Wu, Junkun Chen, Lei Li, Yuan-Fang Wang, and William~Yang Wang. 2020.
\newblock \href {https://arxiv.org/abs/1904.03493} {Vatex: A large-scale, high-quality multilingual dataset for video-and-language research}.
\newblock \emph{Preprint}, arXiv:1904.03493.

\bibitem[{Wang et~al.(2025)Wang, Li, Yan, He, Yu, Zeng, Wang, Ma, Huang, Gao, Dou, Chen, Wang, Qiao, Wang, and Wang}]{wang2025internvideo25empoweringvideomllms}
Yi~Wang, Xinhao Li, Ziang Yan, Yinan He, Jiashuo Yu, Xiangyu Zeng, Chenting Wang, Changlian Ma, Haian Huang, Jianfei Gao, Min Dou, Kai Chen, Wenhai Wang, Yu~Qiao, Yali Wang, and Limin Wang. 2025.
\newblock \href {https://arxiv.org/abs/2501.12386} {Internvideo2.5: Empowering video mllms with long and rich context modeling}.
\newblock \emph{Preprint}, arXiv:2501.12386.

\bibitem[{Wanner et~al.(2024{\natexlab{a}})Wanner, Durme, and Dredze}]{wanner2024dndscoredecontextualizationdecompositionfactuality}
Miriam Wanner, Benjamin~Van Durme, and Mark Dredze. 2024{\natexlab{a}}.
\newblock \href {https://arxiv.org/abs/2412.13175} {Dndscore: Decontextualization and decomposition for factuality verification in long-form text generation}.
\newblock \emph{Preprint}, arXiv:2412.13175.

\bibitem[{Wanner et~al.(2024{\natexlab{b}})Wanner, Ebner, Jiang, Dredze, and Van~Durme}]{wanner-etal-2024-closer}
Miriam Wanner, Seth Ebner, Zhengping Jiang, Mark Dredze, and Benjamin Van~Durme. 2024{\natexlab{b}}.
\newblock \href {https://doi.org/10.18653/v1/2024.starsem-1.13} {A closer look at claim decomposition}.
\newblock In \emph{Proceedings of the 13th Joint Conference on Lexical and Computational Semantics (*SEM 2024)}, pages 153--175, Mexico City, Mexico. Association for Computational Linguistics.

\bibitem[{Weller et~al.(2025)Weller, Ricci, Yang, Yates, Lawrie, and Durme}]{weller2025rank1testtimecomputereranking}
Orion Weller, Kathryn Ricci, Eugene Yang, Andrew Yates, Dawn Lawrie, and Benjamin~Van Durme. 2025.
\newblock \href {https://arxiv.org/abs/2502.18418} {Rank1: Test-time compute for reranking in information retrieval}.
\newblock \emph{Preprint}, arXiv:2502.18418.

\bibitem[{Xu et~al.(2016)Xu, Mei, Yao, and Rui}]{7780940}
Jun Xu, Tao Mei, Ting Yao, and Yong Rui. 2016.
\newblock \href {https://doi.org/10.1109/CVPR.2016.571} {Msr-vtt: A large video description dataset for bridging video and language}.
\newblock In \emph{2016 IEEE Conference on Computer Vision and Pattern Recognition (CVPR)}, pages 5288--5296.

\bibitem[{Yang et~al.(2025)Yang, Chen, Xu, Fei, Shen, Zhao, Feng, and Elhoseiny}]{yang2025wikiautogenmultimodalwikipediastylearticle}
Zhongyu Yang, Jun Chen, Dannong Xu, Junjie Fei, Xiaoqian Shen, Liangbing Zhao, Chun-Mei Feng, and Mohamed Elhoseiny. 2025.
\newblock \href {https://arxiv.org/abs/2503.19065} {Wikiautogen: Towards multi-modal wikipedia-style article generation}.
\newblock \emph{Preprint}, arXiv:2503.19065.

\bibitem[{Yu et~al.(2019)Yu, Xu, Yu, Yu, Zhao, Zhuang, and Tao}]{zhou2019activitynetqa}
Zhou Yu, Dejing Xu, Jun Yu, Ting Yu, Zhou Zhao, Yueting Zhuang, and Dacheng Tao. 2019.
\newblock \href {https://doi.org/10.1609/aaai.v33i01.33019127} {Activitynet-qa: a dataset for understanding complex web videos via question answering}.
\newblock In \emph{Proceedings of the Thirty-Third AAAI Conference on Artificial Intelligence and Thirty-First Innovative Applications of Artificial Intelligence Conference and Ninth AAAI Symposium on Educational Advances in Artificial Intelligence}, AAAI'19/IAAI'19/EAAI'19. AAAI Press.

\bibitem[{Zha et~al.(2023)Zha, Yang, Li, and Hu}]{zha-etal-2023-alignscore}
Yuheng Zha, Yichi Yang, Ruichen Li, and Zhiting Hu. 2023.
\newblock \href {https://doi.org/10.18653/v1/2023.acl-long.634} {{A}lign{S}core: Evaluating factual consistency with a unified alignment function}.
\newblock In \emph{Proceedings of the 61st Annual Meeting of the Association for Computational Linguistics (Volume 1: Long Papers)}, pages 11328--11348, Toronto, Canada. Association for Computational Linguistics.

\bibitem[{Zhang et~al.(2019)Zhang, Kishore, Wu, Weinberger, and Artzi}]{zhang-etal-2019-bertscore}
Tianyi Zhang, Varsha Kishore, Felix Wu, Kilian~Q. Weinberger, and Yoav Artzi. 2019.
\newblock \href {https://api.semanticscholar.org/CorpusID:127986044} {Bertscore: Evaluating text generation with bert}.
\newblock \emph{ArXiv}, abs/1904.09675.

\bibitem[{Zhang et~al.(2024{\natexlab{a}})Zhang, Wu, Li, Li, Ma, Liu, and Li}]{zhang2024videoinstructiontuningsynthetic}
Yuanhan Zhang, Jinming Wu, Wei Li, Bo~Li, Zejun Ma, Ziwei Liu, and Chunyuan Li. 2024{\natexlab{a}}.
\newblock \href {https://arxiv.org/abs/2410.02713} {Video instruction tuning with synthetic data}.
\newblock \emph{Preprint}, arXiv:2410.02713.

\bibitem[{Zhang et~al.(2024{\natexlab{b}})Zhang, Wu, Li, Li, MA, Liu, and Li}]{zhang2024video}
Yuanhan Zhang, Jinming Wu, Wei Li, Bo~Li, Zejun MA, Ziwei Liu, and Chunyuan Li. 2024{\natexlab{b}}.
\newblock \href {https://openreview.net/forum?id=8Livf4oZxz} {Video instruction tuning with synthetic data}.

\bibitem[{Zhou et~al.(2019)Zhou, Kalantidis, Chen, Corso, and Rohrbach}]{Zhou_2019_CVPR}
Luowei Zhou, Yannis Kalantidis, Xinlei Chen, Jason~J. Corso, and Marcus Rohrbach. 2019.
\newblock Grounded video description.
\newblock In \emph{Proceedings of the IEEE/CVF Conference on Computer Vision and Pattern Recognition (CVPR)}.

\bibitem[{Zhu et~al.(2021)Zhu, Tu, Shi, Li, Hou, and Cui}]{zhu-etal-2021-twag}
Fangwei Zhu, Shangqing Tu, Jiaxin Shi, Juanzi Li, Lei Hou, and Tong Cui. 2021.
\newblock \href {https://doi.org/10.18653/v1/2021.acl-long.356} {{TWAG}: A topic-guided {W}ikipedia abstract generator}.
\newblock In \emph{Proceedings of the 59th Annual Meeting of the Association for Computational Linguistics and the 11th International Joint Conference on Natural Language Processing (Volume 1: Long Papers)}, pages 4623--4635, Online. Association for Computational Linguistics.

\end{thebibliography}

\appendix

\appendix
\section{Dataset Statistics}
\label{append:stats}

\autoref{tab:claim_stats} contains additional statistics about the claims in \dsetname  and \autoref{tab:video_stats} has additional statistics about the videos. Further examples of \dsetname articles can be found in \autoref{tab:model_human_ref1} and \autoref{tab:model_human_ref2}. We also report the IAA metrics in \autoref{tab:iaa}.

% \begin{table}[]
%     \centering
%     \begin{tabular}{c|ccccccc}
%          &  ar & en & es & ko & ru & zh & Total \\
%          \midrule
%          Topics & 24 & 43 & & 31 & 28 & 31 & 151 \\
%          Videos & 461 & 504 & 2 & 528 & 497 & 537 & 2524  \\
%          % \midrule
%          % Audio Claims & - & - & - & - & - & - & - \\
%          % Video Claims & - & - & - & - & - & - & - \\ 
%     \end{tabular}
%     \caption{Topic Counts and Language Splits. }
%     \label{tab:my_label}
% \end{table}
% Claims found in audio: 1299
% Claims found in video: 954
% Claims found in both: 674
% Claims not found: 24127
% Total claims: 26782
% Average kept claims per topic: 51.05769230769231
% Average audio claims per topic: 24.98076923076923
% Average video claims per topic: 18.346153846153847
% Average claims found in both: 12.961538461538462
% Max videos: 12
% Min videos: 1
% Average videos: 7.653846153846154
% Total number of videos: 398
\begin{table}[ht]
    \centering
    \begin{tabular}{l|r}
         \toprule
            Video-Grounded Claims & 954 \\
            Audio-Grounded Claims & 1299 \\
            A+V-Grounded Claims & 674 \\
            Avg.\ Claims / Event & 51.10 \\
            Avg.\ Audio Claims / Event & 24.98 \\
            Avg.\ Video Claims / Event & 18.35 \\
            Avg.\ OCR Claims / Event & 28.53 \\
            Avg.\ Claims in All / Event & 12.96 \\
        \bottomrule
    \end{tabular}
    \caption{Claim Statistics}
        \label{tab:claim_stats}
\end{table}

\begin{table}[ht]
    \centering
    \begin{tabular}{l|r}
         \toprule
            Max Videos for a Topic & 12 \\
            % Min Videos for a Topic & 1 \\ 
            Avg.\ Videos / Event & 7.49 \\
            Total Relevant Videos & 398 \\
            RAG Video Data Lake & 109K \\
            Avg.\ Video Length (Relevant) & 79.57s \\
            Avg.\ Video Length (RAG) & 145s \\
            Max Relevant Video Length & 586.26s \\
            Min Relevant Video Length & 4.55s \\
        \bottomrule
    \end{tabular}
    \caption{Video Statistics}
    \label{tab:video_stats}
\end{table}

\begin{table}[]
    \centering
    \begin{tabular}{c|c}
    \toprule
        \textbf{Modality} & \textbf{$\alpha$} \\
        \midrule
         Video & 0.446 \\
         Audio  & 0.780 \\
         OCR  & 0.722 \\
         None  & 0.682 \\
         Overall  & 0.767 \\
    \bottomrule
    \end{tabular}
    \caption{Krippendorff's $\alpha$ for the claim grounding agreement of each modality. Each judgment reflects a binary decision about whether a given claim is supported by a given modality (or \textbf{None} of the modalities).}
    \label{tab:iaa}
\end{table}

% We additionally provide example articles for videos in the MultiVENT-G~\citep{sanders-etal-2024-grounding} event ontology of Elections (\autoref{fig:election_example}), Political Developments (\autoref{fig:political_dev_example}), Social Events (\autoref{fig:social_example}), Sports (\autoref{fig:sports_example}), disocoveries/launches (\autoref{fig:disc_launch_example}), and Demonstrations (\autoref{fig:demon_example}).

\section{Data Collection}
\label{append:annotation}

This appendix discusses the annotation process for \dsetname in greater detail. Beyond topic selection, recall that the annotation process includes claim decomposition, subclaim rewriting, subclaim grounding, and article rewriting.

\subsection{Claim Decomposition}
\label{append:decomp}
Much recent work has studied the appropriate granularity of subclaims in a claim decomposition and the applications of such decompositions to natural language inference (NLI) and claim verification \citep{min-etal-2023-factscore, gunjal-durrett-2024-molecular, wanner-etal-2024-closer, wanner2024dndscoredecontextualizationdecompositionfactuality, hu2025decompositiondilemmasdoesclaim, srikanth2025nlimicroscopeatomichypothesis}. Our own claim decomposition method is most similar to those of \cite{gunjal-durrett-2024-molecular} and \cite{wanner2024dndscoredecontextualizationdecompositionfactuality} in that we \emph{decontextualize} subclaims---insert elided or abstracted context (e.g.\ by substituting pronouns with named entities)---up to the point that no further verification of extracted facts is required. (An example of one of our claim decomposition prompts can be found in \autoref{prompt:claim_decomp}). % However, we do not go as far as \cite{gunjal-durrett-2024-molecular} (see their Figure 1), to decontextualize to the extent that it is verifiable against a knowledge base. For example: ``Ann Jansson, a
% Swedish footballer, won a medal at the
% European Athletics Championships.'' includes the additional claim ``Swedish footballer'' that we would not include, instead decomposing as ``Ann Jansson won a medal...'' and ``Ann Jansson is a Swedish footballer.''
Although not a rigourous form of decomposition, this notion is straightforward to apply during annotation. Additionally for simplicity, we treat dates as named entities and do not decompose them beyond their mention as given in the text. For example, some methods, like \cite{wanner-etal-2024-closer}, may decompose the claim ``The event occurred on 15 April 2019'' as:
\begin{itemize}
    \item The event occurred on the 15th
    \item The event occurred in April
    \item The event occurred in 2019
\end{itemize}
However, this produces an additional burden on our downstream annotations requiring the annotator to verify all 3 concepts. While it is possible to fail to recover finer-grained information that may be attested in the video---e.g. \emph{the event occurred in 2019}---such cases are rare and this decision substantially reduces the amount of labor required for subclaim rewriting and grounding.

% To further strengthen this stance, we believe that it is not the job of the annotation process to have 100\% recall (although the process should have 100\% precision). Instead, recall should be considered in the downstream evaluation. This is prevelant in computer vision cases like image captioning. Take COCO \citep{lin2015microsoftcococommonobjects} as an example. These images can still be used for evaluation if the evaluation leaves room for the fact that caption is non-exhaustive for every visual concept possible in an image and VLMs are often more expressive than these captions. An example of such metric: FaithScore \citep{jing-etal-2024-faithscore}.

\begin{figure*}
\noindent\fbox{%
    \parbox{.98\textwidth}{%
% \textbf{Few-shot Prompt for Claim Decomposition}\\

\footnotesize
{\tt
\tiny
Instructions:
- You are given a paragraph, and one sentence from the paragraph to decompose
- You must decompose this into a set of claims
- You must decompose this into a JSON format [{"claim": "..."}, {"claim": "..."}, ...]

PARAGRAPH: On 15 April 2019, just before 18:20 CEST, a structural fire broke out in the roof space of Notre-Dame de Paris, a medieval Catholic cathedral in Paris, France. By the time the fire was extinguished, the cathedral\"s wooden spire (flèche) had collapsed, most of the wooden roof had been destroyed, and the cathedral\"s upper walls were severely damaged. Extensive damage to the interior was prevented by the vaulted stone ceiling, which largely contained the burning roof as it collapsed. Many works of art and religious relics were moved to safety, but others suffered smoke damage, and some of the exterior art was damaged or destroyed. The cathedral\"s altar, two pipe organs, and three 13th-century rose windows suffered little or no damage. Three emergency workers were injured. The fire contaminated the site and nearby areas of Paris with toxic dust and lead. Notre-Dame did not hold a Christmas Mass in 2019, for the first time since 1803. Investigators in 2020 believed the fire to have been "started by either a cigarette or a short circuit in the electrical system".

SENTENCE: On 15 April 2019, just before 18:20 CEST, a structural fire broke out in the roof space of Notre-Dame de Paris, a medieval Catholic cathedral in Paris, France.

DECOMPOSITION: 
[
    {"claim": "A structural fire broke out"}, 
    {"claim": "The fire broke out on 15 April 2019"}, 
    {"claim": "The fire broke out just before 18:20 CEST"}, 
    {"claim": "The fire broke out in the roof space"}, 
    {"claim": "Notre-Dame de Paris is a medieval Catholic cathedral"}, 
    {"claim": "Notre-Dame de Paris is located in Paris, France"}
]

PARAGRAPH: The 2022 United States Senate election in Georgia was held on November 8, 2022, to elect a member of the U.S. Senate to represent the state of Georgia. Incumbent Democratic senator Raphael Warnock won his first full term in office, defeating Republican former football player Herschel Walker. Under Georgia"s two-round system, Warnock was re-elected in a runoff election on December 6 after neither candidate received over 50\% of the vote on November 8. Warnock"s win was the only statewide victory for Democrats in Georgia in 2022, as Republicans swept all other races.

SENTENCE: Under Georgia"s two-round system, Warnock was re-elected in a runoff election on December 6 after neither candidate received over 50% of the vote on November 8.

DECOMPOSITION: 
[
    {"claim": "Georgia has a two-round election system"}, 
    {"claim": "The runoff election is part of the two-round election system in Georgia"}, 
    {"claim": "A run off election occurs when no candidate receives 50\% of the vote"}, 
    {"claim": "Neither candidate received more than 50\% of the vote in the first election"}, 
    {"claim": "The runoff election took place on December 6"}, 
    {"claim": "Warnock won the runoff election"}, 
    {"claim": "Warnock was the incumbent candidate"}, 
    {"claim": "The first election took place on November 8"}
]

PARAGRAPH: Hurricane Irma was an extremely powerful Cape Verde hurricane that caused widespread destruction across its path in early September 2017. Irma was the first Category 5 hurricane to strike the Leeward Islands on record, followed by Maria two weeks later. At the time, it was considered the most powerful hurricane on record in the open Atlantic region, outside of the Caribbean Sea and Gulf of Mexico, until it was surpassed by Hurricane Dorian two years later. It was also the third-strongest Atlantic hurricane at landfall ever recorded, just behind the 1935 Labor Day Hurricane and Dorian. The ninth named storm, fourth hurricane, second major hurricane, and first Category 5 hurricane of the extremely active 2017 Atlantic hurricane season, Irma caused widespread and catastrophic damage throughout its long lifetime, particularly in the northeastern Caribbean and the Florida Keys. It was also the most intense hurricane to strike the continental United States since Katrina in 2005, the first major hurricane to make landfall in Florida since Wilma in the same year, and the first Category 4 hurricane to strike the state since Charley in 2004. The word Irmageddon was coined soon after the hurricane to describe the damage caused by the hurricane.

SENTENCE: The ninth named storm, fourth hurricane, second major hurricane, and first Category 5 hurricane of the extremely active 2017 Atlantic hurricane season, Irma caused widespread and catastrophic damage throughout its long lifetime, particularly in the northeastern Caribbean and the Florida Keys.

DECOMPOSITION:
[
    {"claim": "Irma was the ninth named storm of the 2017 Atlantic hurricane season"}, 
    {"claim": "Irma was the fourth hurricane of the 2017 Atlantic hurricane season"}, 
    {"claim": "Irma was the second major hurricane of the 2017 Atlantic hurricane season"}, 
    {"claim": "Irma was the first Category 5 hurricane of the 2017 Atlantic hurricane season"}, 
    {"claim": "The  2017 Atlantic hurricane season was extremely active"}, 
    {"claim": "Irma caused widespread damage"}, 
    {"claim": "Irma caused catastrophic damage"}, 
    {"claim": "Irma's damage was particularly severe in the northeastern Caribbean"}, 
    {"claim": "Irma's damage was particularly severe in the Florida Keys"}, 
    {"claim": "Irma had a long lifetime"}, 
    {"claim": "Irma occurred during the 2017 Atlantic hurricane season"}
]

PARAGRAPH: On November 30, 2018, at 8:29 a.m. AKST (17:29 UTC), a magnitude 7.1 earthquake hit Anchorage in South Central Alaska. The earthquake's epicenter was near Point Mackenzie, about north of Anchorage, and occurred at a depth of . It was followed six minutes later by a magnitude 5.7 aftershock centered north-northwest of the municipality. The earthquake could be felt as far away as Fairbanks.

SENTENCE: On November 30, 2018, at 8:29 a.m. AKST (17:29 UTC), a magnitude 7.1 earthquake hit Anchorage in South Central Alaska.

DECOMPOSITION:
[
    {"claim": "The earthquake occurred"}, 
    {"claim": "The earthquake occurred on November 30, 2018"}, 
    {"claim": "The earthquake occurred at 8:29 a.m. AKST"}, 
    {"claim": "The earthquake hit Anchorage"}, 
    {"claim": "The earthquake hit South Central Alaska"}, 
    {"claim": "The earthquake had a magnitude of 7.1"}
]

PARAGRAPH: Pokémon Go (stylized as Pokémon GO) is a 2016 augmented reality (AR) mobile game, part of the Pokémon franchise, developed and published by Niantic in collaboration with Nintendo and The Pokémon Company for iOS and Android devices. It uses mobile devices with GPS to locate, capture, train, and battle virtual Pokémon, which appear as if they are in the player's real-world location. The game is free-to-play; it uses a freemium business model combined with local advertising and supports in-app purchases for additional in-game items. The game launched with around 150 species of Pokémon, which had increased to around 700 by 2021.

SENTENCE: Pokémon Go (stylized as Pokémon GO) is a 2016 augmented reality (AR) mobile game, part of the Pokémon franchise, developed and published by Niantic in collaboration with Nintendo and The Pokémon Company for iOS and Android devices.

DECOMPOSITION:
[
    {"claim": "Pokémon Go is a mobile game"}, 
    {"claim": "Pokémon Go is an augmented reality (AR) game"}, 
    {"claim": "Pokémon Go was released in 2016"}, 
    {"claim": "Pokémon Go is part of the Pokémon franchise"}, 
    {"claim": "Pokémon Go was developed by Niantic"}, 
    {"claim": "Pokémon Go was published by Niantic"}, 
    {"claim": "Niantic collaborated with Nintendo to develop Pokémon Go"}, 
    {"claim": "Niantic collaborated with Nintendo to publish Pokémon Go"},
    {"claim": "Niantic collaborated with The Pokémon Company to develop Pokémon Go"}, 
    {"claim": "Niantic collaborated with The Pokémon Company to publish Pokémon Go"}, 
    {"claim": "Pokémon Go was developed for iOS devices"}, 
    {"claim": "Pokémon Go was published for iOS devices"}, 
    {"claim": "Pokémon Go was developed for Android devices"}, 
    {"claim": "Pokémon Go was published for Android devices"}
]

PARAGRAPH [paragraph]
SENTENCE [sentence]
DECOMPOSITION:
}
}}

\caption{Prompt For Qwen 2.5 32B Claim Decomposition}
\label{prompt:claim_decomp}
\end{figure*}
\subsection{Subclaim Rewriting}
\label{append:claim_rewrite}
% While we take the above stance in \autoref{append:decomp} that the process should not have 100\% recall, we do want to ensure the highest possible recall. To do this,

\begin{figure*}
    \centering
    \includegraphics[width=\linewidth]{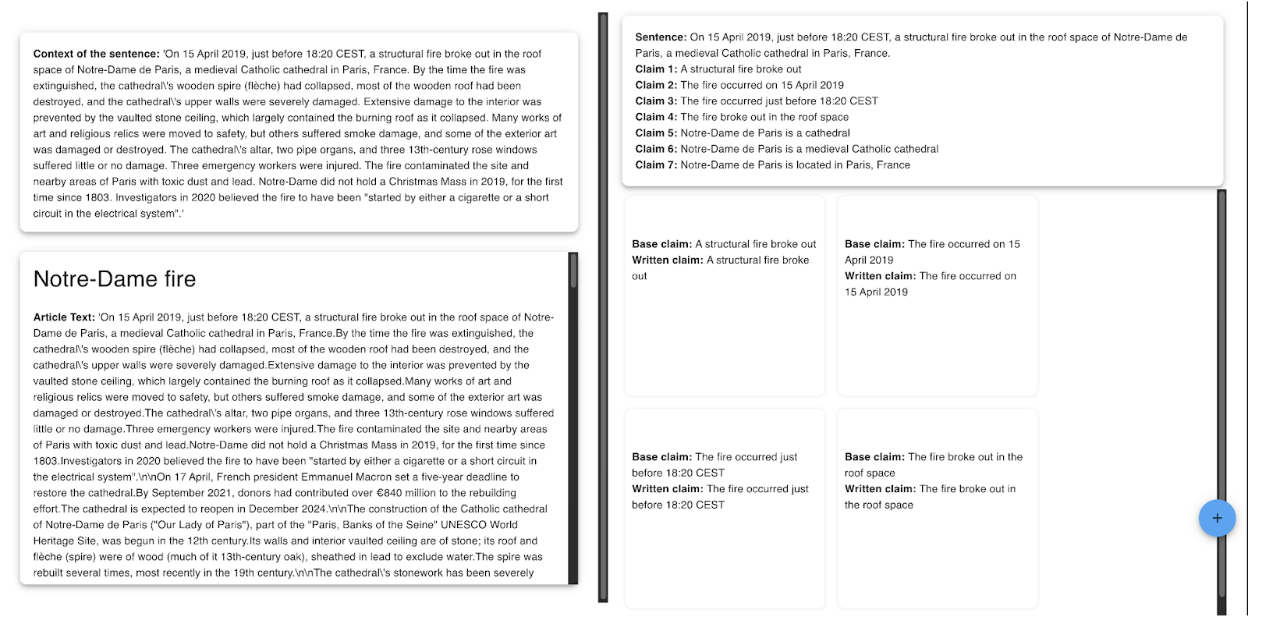}
    \caption{The annotation interface for our subclaim grounding task. In this protocol, the left hand side is both versions of the Wikipedia context. The top context is the paragraph a sentence comes from and the bottom context is the lead section of the Wikipedia article. On the right hand side is the sentence to be decomposed and its claims. The claims from Qwen32B are prepopulated in the protocol and the rewriters edit them.}
    \label{fig:claim-rewrite-protocol}
\end{figure*}

We take the subclaims decomposed by Qwen2.5 32B and correct them manually, with three of the authors serving as annotators. \autoref{fig:claim-rewrite-protocol} shows the interface that the annotators used to rewrite, add, and remove claims, and \autoref{prompt:claim_rewrite_instrut} shows the instructions provided to annotators.

\begin{figure*}
\noindent\fbox{%
    \parbox{.98\textwidth}{%
% \textbf{Few-shot Prompt for Claim Decomposition}\\

\footnotesize
{\tt
\tiny
Claim Rewriting Annotation 

In this task, you will be shown an excerpt from Wikipedia, a sentence, and a set of claims associated with that sentence. Given this information, you will rewrite the set of claims into a new set of “cleaner” claims. To create the final set of claims, you will deal with common issues like splitting a claim into two or more claims, adding a missing claim, or removing a duplicated claim. 

Do not remove duplicates you remember from a previous sentence of the document. Treat each sentence as a unique instance without recalling what you had annotated previously. 

Annotation Protocol

The annotation protocol has 4 main sections. The R.H.S. is what is most relevant to annotators. It contains the current sentence and the claims for the sentences as well as editable cards with each claim from the sentence in them. (Sometimes not all of them are prepopulated, so if any are missing just hit the + button). The L.H.S. has the context that the sentences are taken from. The top context is the paragraph where the current sentence comes from and below is the lead section of the wikipedia article that the paragraph comes from. 

Editing a claim

To edit a claim, you will click on the card that the claim is in. The claim will include above it the original claim and the textbox will allow you to delete, edit, and rewrite claims. 

Adding extra claims
Sometimes claims are missed by the decomposer. To fix these press the + button(   ) in the right side of the interface. This will put a new claim into the interface. 
Note that these claims won’t include base claims in the interface because the system did not predict the claim. This will NOT impact the annotation or future use of the claim.

Error Types
Here are some examples of errors that you might encounter when doing the annotations.

Under Decomposition 
Under decomposition is when a claim includes multiple pieces of information that could be split into two or more atomic facts. 

Under Decomposition Example (loaded claim):
Input: On 15 April 2019, just before 18:20 CEST, a structural fire broke out in the roof space of Notre-Dame de Paris, a medieval Catholic cathedral in Paris, France.
Incorrect Decomposition: 
A structural fire broke out
The fire occurred on 15 April 2019
The fire occurred just before 18:20 CEST
The fire broke out in the roof space
Notre-Dame de Paris is a cathedral
Notre-Dame de Paris is a medieval Catholic cathedral
Notre-Dame de Paris is located in Paris, France
Correct Decomposition:
A structural fire broke out
The fire occurred on 15 April 2019
The fire occurred just before 18:20 CEST
The fire broke out in the roof space
Notre-Dame de Paris is a cathedral
Notre-Dame de Paris is a medieval Catholic cathedral
Notre-Dame de Paris is located in Paris
Notre-Dame de Paris is located in France
Reasoning: Original claim 7 had 2 pieces of information in it: The location of Paris and The location of France. While this may seem intuitive that Paris is in France, from an evaluation perspective, it is better to have two distinct claims to verify. See another brief example below:

Incorrect Decomposition:
The event happened in Seoul, South Korea
Correct Decomposition: 
The event happened in Seoul
The event happened in South Korea
Reasoning: When considering the evaluation of where the event happened, it’s better to split the claim into the two locations: Seoul and South Korea, so that you can evaluate against systems that say only South Korea (if it happened in other locations in SK) or against systems that only state the city.

Hallucinated Decomposition
This is the addition of a claim that isn’t supported by the sentence. 

Input: The 2016 World Short Track Speed Skating Championships took place from 11 to 13 March 2016 in Seoul, South Korea.
Incorrect Decomposition:
The event took place
The event was from 11 to 13 March 2016
The event happened in Seoul
The event happened in South Korea
The event was the 41st speed skating championship
Correct Decomposition:
The event took place
The event was from 11 to 13 March 2016
The event happened in Seoul
The event happened in South Korea
Reasoning: 5 is factual, but it is not supported by the sentence.

Ambiguity
Only resolve ambiguity if the claims make it difficult to disambiguate between entities. 

Input: The 2016 World Short Track Speed Skating Championships took place from 11 to 13 March 2016 in Seoul, South Korea.
Correct Decomposition:
The event took place
The event was from 11 to 13 March 2016
The event happened in Seoul
The event happened in South Korea
Reasoning: Only one event in the sentence. No need to disambiguate. 

Input: Due to Imran Khan’s criticism of Macron’s comments on Islam, French authorities cancelled the visas of 183 Pakistani citizens and deported 118 from the country.
Incorrect Decomposition:
They cancelled the visas of 183 Pakistani citizens.
They deported 118 Pakistani citizens from the country.
He criticized Macron’s comments on Islam
Correct Decomposition: 
French authorities cancelled the visas of 183 Pakistani citizens.
French authorities deported 118 Pakistani citizens from the country.
Imran Khan criticized Macron’s comments on Islam
Reasoning: In this scenario, it’s better to disambiguate the references to the named entities because they could be Imran, Macron, or the French authorities.

Current notes / edge cases during test evaluation
Go over 2022 Senate Election annotations.

Sentence: The National Tsunami Warning Center—itself located inside the quake zone, in Palmer, Alaska, northeast of Anchorage—issued tsunami warnings for nearby coastal areas, including Cook Inlet and the Kenai Peninsula, but they were lifted shortly after.
Claim 1: The National Tsunami Warning Center issued warnings
Claim 2: The warnings were for nearby coastal areas
Claim 3: The warnings included Cook Inlet
Claim 4: The warnings included the Kenai Peninsula
Claim 5: The warnings were lifted shortly after issuance
Claim 6: Palmer is located in Alaska
Claim 7: Palmer is northeast of Anchorage
Claim 8: The National Tsunami Warning Center is located in Palmer
Claim 9: The National Tsunami Warning Center is inside the quake zone
Claim 10: Cook Inlet is a coastal area
Claim 11: The Kenai Peninsula is a coastal area

6,7,8 as claims related to the location of NTWC or Palmer. Perspective matters probably. I would may rewrite these to be all about the NTWC location.

Sentence: Notre-Dame did not hold a Christmas Mass in 2019, for the first time since 1803.
Claim 1: Notre-Dame did not hold a Christmas Mass in 2019
Claim 2: Notre-Dame did not hold a Christmas Mass was in 1803
Claim 3: Notre-Dame held a Christmas Mass every year between 1803 and 2019

Sentence: Investigators in 2020 believed the fire to have been "started by either a cigarette or a short circuit in the electrical system".
Claim 1: The investigators believed the fire was started
Claim 2: The investigators identified two possible causes for the fire
Claim 3: One possible cause was a cigarette
Claim 4: Another possible cause was a short circuit in the electrical system
Claim 5: The investigation took place in 2020

New:
There was an investigation into the cause of the fire
The investigators identified two possible causes for the fire
The fire was possibly started by a cigarette
The fire was possibly started by a short circuit in the electrical system

The investigation took place in 2020 (this might not be factual)
How to incorporate the date? 
In 2020 investigators believed the fire was started.

original: 5
['The investigators believed the fire was started', 'The investigators identified two possible causes for the fire', 'One possible cause was a cigarette', 'Another possible cause was a short circuit in the electrical system', 'The investigation took place in 2020']
2958: 4
['The investigators identified two possible causes for the fire', 'A possible cause was a cigarette', 'A possible cause was a short circuit in the electrical system', 'The investigation took place in 2020']
2959: 6
['The investigators believed the fire was started', 'The investigators identified two possible causes for the fire', 'The fire was possibly started by a cigarette', 'The fire was possibly started by a short circuit.', 'The investigation took place in 2020', 'The possible short circuit occurred in the electrical system']
2960: 5
['There was an investigation into the cause of the fire.', 'Investigators identified two possible causes for the fire', 'One possible cause was a cigarette', 'One possible cause was a short circuit in the electrical system', 'An investigation took place in 2020']
}
}}

\caption{Annotation Instructions for Claim Rewriting}
\label{prompt:claim_rewrite_instrut}
\end{figure*}
\subsection{Subclaim Grounding}
\label{append:grounding}
\begin{figure*}
    \centering
    \includegraphics[width=\linewidth]{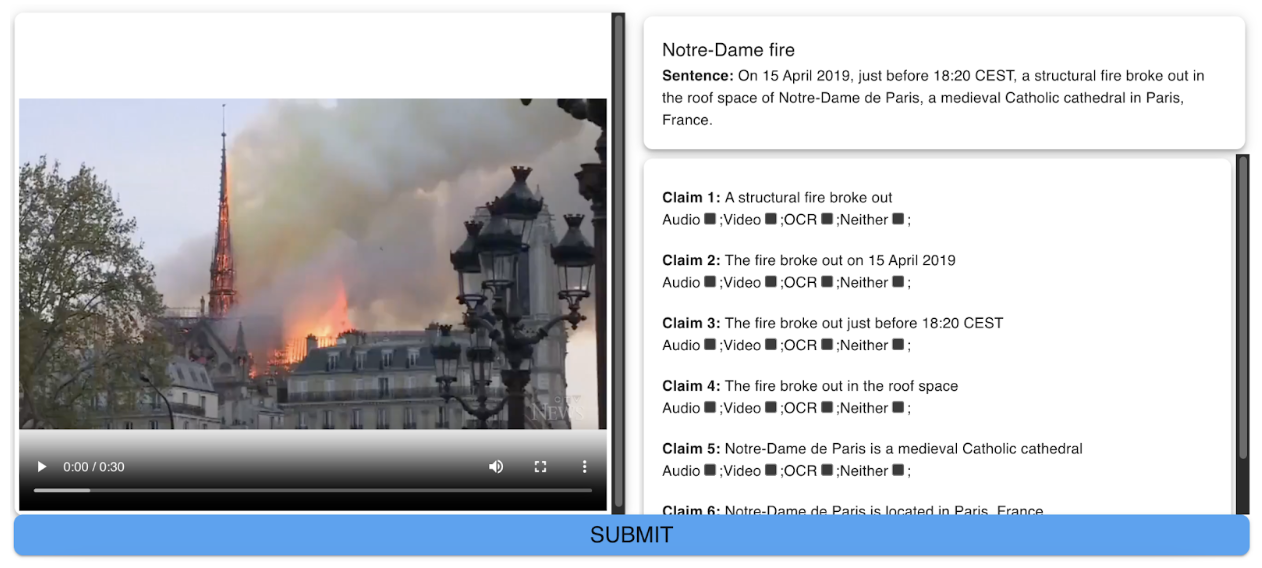}
    \caption{The claim grounding protocol. This protocol has the video on the left hand side and a sentence and its claims on the right. Each claim has 4 buttons which the annotator can select for the modality (or none) that support the claim.}
    \label{fig:grounding-protocol}
\end{figure*}

Once the final set of claims is completed by the human annotator, the same annotators ground the claims in the video content. We provide the instructions for this process in \autoref{prompt:claim_grounding_instruct} and the protocol for this populated with an example sentence worth of claims and video from the Notre-Dame fire in \autoref{fig:grounding-protocol}. 

Note that this grounding annotation was initially attempted via Amazon Mechanical Turk. However, despite several iterations of annotation instructions, even Turkers with additional Masters qualifications struggled to consistently ground claims in videos. We suspect this is in part due to the amount of domain expertise and world knowledge required to properly ground claims, and partially because raw real-time videos of events are inherently ambiguous. Thus, the decision was made to leverage domain experts in order to ensure consistent and high-quality annotations.

% The primary author would also like to express how much they hate Amazon Mechanical Turk after this experience. The author tried to use Turkers with the \textbf{Masters} qualification and spent so long on task instructions and qualifying workers to still have the task be poorly annotated. 

\begin{figure*}
\noindent\fbox{%
    \parbox{.98\textwidth}{%
% \textbf{Few-shot Prompt for Claim Decomposition}\\

\footnotesize
{\tt
\tiny
This task involves watching a video and deciding whether the video supports particular claims. A video supports a claim if it can be verified from the video’s visual, audio, or text content.

Visual text (denoted “Text” in the interface and instructions) support refers to on-screen text that supports the claim. This may include (e.g.) text on a street sign, text scrolling across the screen in a news broadcast, text on someone's clothing, subtitles on the screen, etc.

Visual support refers to anything else on-screen besides text that supports the claim. This includes any action happening on-screen, still images, or other graphics (e.g. a map shown in a weather report) or animations.

Audio support refers to any sounds (e.g. sirens, gunshots) or speech (e.g. from a newscaster or from the person filming) that supports the claim.

Each claim may have only one of these types of support in a given clip, may have multiple types of support, or may not be supported at all. It is your job to determine which type(s) of support there are for each claim.

Note: Sometimes the clips may contain audio (speech) or visual text in a language that you do not speak. For these instances, do not try to annotate claims for the information in another language. 
Task Interface
On the left side of the interface, you will see a video clip.

You can adjust the playback speed by clicking on the three vertical dots at the bottom right of the clip:

You can adjust the video’s playback speed by clicking the three vertical dots on the bottom right:

Note: We do not suggest making the video full screen.

On the right side of the interface, you will see:
The name of the event that the clip is about
a sentence about that event
a list of claims derived from that sentence

Each claim comes with three buttons: Audio (for audio support), Visual (for visual support), Text (for text support), and Neither (for none of the above):

For each claim, you should check all boxes corresponding to the types of support for the claim attested in the video:

If a claim has no support in the video, you should click only the Neither checkbox:

You are allowed to select any combination of the Audio, Visual, and Text checkboxes, but the interface will prevent you from selecting the Neither checkbox in combination with any of the other three.

When all claims have at least one button checked, the HIT will become ready for submission: the SUBMIT button will change from gray to blue and you then can submit the HIT:

Task Instructions
Your job is to:

Watch the video clip
For each claim, select all checkboxes corresponding to the type(s) of support it receives in the video (Audio, Text, or Visual) or Neither if it has no support

Note that you are allowed—and encouraged!—to rewatch the video if necessary in order to assess the types of support it provides for a particular claim. At least one checkbox must be clicked for each claim before you can submit the HIT.
What does it mean for a claim to be supported by visuals/audio/text?
You can conceive of each claim as both posing a question and providing an answer to that question.

For example, the claim Dominguez hit a home run can be understood to be:
Implicitly asking the question: Did Dominguez hit a home run?
Answering this question affirmatively: Yes, he did.

If the audio (sounds, speech), visuals (action, images, graphics), and on-screen text provides the same answer to the question as the claim—i.e. if the audio, visuals, or text provide evidence that the claim is true—then you should click the corresponding checkbox.

Alternatively, the clip may not provide an answer to the claim’s question at all, or may even provide an answer that contradicts the answer implicit in the claim. In these cases, you should select only the Neither checkbox.

Note: in many cases, it will be difficult to say with absolute certainty whether a given claim is supported or not by the audio, visuals, or text. The standard for determining whether a claim has one of these types of support is not certainty, but rather high confidence, given the clip’s contents. You will have to rely on your own judgment of what a normal person could confidently infer about the truth of the claim, having watched the video yourself. (See World Knowledge Support)

Audio Support Examples
Audio support may come from speech or other sounds in the video clip. Below are two examples of cases in which a claim has audio support.

Example 1

Claim: Dominguez hit the first home run of his career.
Video: https://www.youtube.com/watch?v=c6U4AnW4ohM
Audio Evidence: The broadcaster in the video announces the homerun that Dominguez hit.

Example 2

Claim: Emergency responders went to the scene
Video: https://www.youtube.com/watch?v=rVKwa4ZqQAA
Audio Evidence: You can hear the sirens of the fire trucks in this video. Thus, audio support. 
Visual Support Examples
Visual support can come from any type of non-text visual content, including any action happening in the video clip, or still images or graphics that are shown. We include two examples of visual support below.

Example 3

Claim: Dominguez hit a homerun
Video: https://www.youtube.com/watch?v=c6U4AnW4ohM
Explanation: In the video you can see Dominguez hit the homerun because he swings, hits the ball, it goes into the stands. 

Example 4

Claim: The event took place in France
Video: (shows the image of the Eiffel Tower below)

Explanation: The Eiffel Tower is an iconic landmark in Paris, France. Even if the video doesn’t explicitly state that it is located in France, you can very confidently infer that the location is France based on this knowledge that the Eiffel Tower is in France.

Example 5

Claim: The hurricane hit florida
Video: (shows the graphic of the Hurricane’s trajectory below)

Explanation: Although the map doesn’t explicitly say “Florida” you can see from the map that the Hurricane’s path goes through Miami (and thus hits Flordia) 
Text Support Examples
Text support can come from any type of text visible on screen. This can be text deployed on the screen in a news broadcast, street signs, posters in a protest, etc. Some examples of text support are shown below.

Example 6

Claim: The event took place on August 30th.
Video: (includes the frame below)
Explanation: The date is clearly displayed on the bottom left of the screen (30 Aug)

Example 7

Claim: City of London police were deployed.
Video: (includes the frame below)
Explanation: It’s clear from the text on the officer’s back in the frame below that the people shown are City of London police, who appear to be deployed.

Example 8

Claim: Mueller subpoenaed a former aide of Trump.
Video: (includes the frame below)
Explanation: The text shown in the “Breaking News” banner at the bottom of the screen clearly supports the claim.

World Knowledge Support

Claim: Texas borders Mexico 

Explanation: Here in this example, nothing on the map shows the names of Mexico or Texas, but using common knowledge, you can see both the outline of Texas and Mexico. Thus,

Neither Support Examples
Claim: Dominguez hit the last home run of his career
Video: https://www.youtube.com/watch?v=c6U4AnW4ohM
Explanation: This is about his first ever home run. You cannot tell that his career has ended or that he hasn’t hit a homerun since this game.
}
}}

\caption{Instructions for the Claim Grounding Protocol. }
\label{prompt:claim_grounding_instruct}
\end{figure*}
\subsection{Article Rewriting}
\label{append:article_rewrite}
The last stage in the annotation process involves rewriting the original Wikipedia articles based on the grounded claims. The instructions for this are provided in \autoref{prompt:rewrite}.

\begin{figure*}[ht]
\noindent\fbox{%
    \parbox{.98\textwidth}{%
% \textbf{Few-shot Prompt for Claim Decomposition}\\

\footnotesize
{\tt
\tiny
Your task is to write a new Wikipedia article to exclude the claims not found in the video content. You will be given a set of claims and the sentences they come from on the L.H.S. of the protocol and your job will be to rewrite the article / sentences such that only the supported claims are presented in the article. You should try to diversify your writing from the Wikipedia if possible without stepping too far away from the general “Wikipedia” style. 

}
}}

\caption{Instructions for Article Rewriting}
\label{prompt:rewrite}
\end{figure*}

\section{Relevance and RePrompt}
\label{append:reprompter}
In this section, we include the prompt used with the reasoning model (\autoref{prompt:reasoner_prompt}) and examples of real queries provided to the VideoLLM (\autoref{prompt:reprompts}). We note that we had to add force the model to produce the prefix "Describe the video in detail and focus on ..." or else sometimes the reasoner would not follow our instructions and produce a new query. 

\begin{figure*}[ht]
\noindent\fbox{%
    \parbox{.98\textwidth}{%
% \textbf{Few-shot Prompt for Claim Decomposition}\\

\footnotesize
{\tt
\tiny
I am trying to find information about EVENT\_QUERY. I will show you a video summary that might be related to the event. Based on the current summary, can you think of a new query that might help me find more information about the event? Please write a new query that you think will help me find more information about the event. DO NOT write anything except for the new query. If you think the current summary is sufficient, you can say 'no new query.' Otherwise, start your new query with 'Describe the video in detail and focus on' Here is the video summary:
}
}}

\caption{Prompt For Qwen32B Distilled R1.}
\label{prompt:reasoner_prompt}
\end{figure*}

\begin{figure*}[ht]
\noindent\fbox{%
    \parbox{.98\textwidth}{%
% \textbf{Few-shot Prompt for Claim Decomposition}\\

\footnotesize
{\tt
\tiny
\begin{enumerate}
    \item "Describe the video in detail and focus on the specific examples of CRISPR applications in medicine and agriculture mentioned, as well as the ethical considerations discussed."
    \item "Describe the video in detail and focus on the specific demands of the protesters and any notable incidents or interactions during the convoy."
    \item "Describe the video in detail and focus on the specific locations affected, the extent of damage caused, and any unique geological features observed during the eruption."
    \item "Describe the video in detail and focus on the eruption's causes and effects in relation to the earthquake and tsunami."
    \item "Describe the video in detail and focus on the specific mission details, such as the mission name, duration, objectives, and any unique features of the spacecraft or crew."
\end{enumerate}
}
}}

\caption{RePrompts from R1 Provided to a VideoLLM}
\label{prompt:reprompts}
\end{figure*}

\section{Article Generation}
\label{append:article_gen}
% \alex{put table showing stability of method (or instability) across different Qwen LLM sizes.}

% \input{tables/extra_main_results}
% \input{tables/extra_modality_results}

In this section we present the prompt for article synthesis (\autoref{prompt:article_gen})

\begin{figure*}[ht]
\noindent\fbox{%
    \parbox{.98\textwidth}{%
% \textbf{Few-shot Prompt for Claim Decomposition}\\

\footnotesize
{\tt
\tiny
You are an experienced Wikipedia editor. You will be shown summaries of one or more videos related to the same event. Your task is to write the lead section of a Wikipedia article based ONLY on the information provided in the video summary or summaries. The lead section MUST match the quality, style, and tone of real Wikipedia articles. DO NOT write in the style of a news journalist. DO NOT use any external sources or additional knowledge you have about the event. DO NOT output anything other than the Wikipedia lead section. DO NOT refer to any of the videos explicitly in your output. DO NOT write anything except for the Wikipedia lead section, even if the summaries are cut off. You MUST start your output with "<lead>". ONLY START YOUR REPORT WITH <lead>. DO NOT WRITE ANYTHING EXCEPT FOR THE WIKIPEDIA ARTICLE.
}
}}

\caption{Prompt For Article Synthesis.}
\label{prompt:article_gen}
\end{figure*}

\section{Qualitative Differences Between Outputs}
\label{append:outputs}
\begin{table*}[]
    \centering
    \begin{tabular}{|p{0.16\linewidth}|p{0.16\linewidth}|p{0.16\linewidth}|p{0.16\linewidth}|p{0.16\linewidth}|}
         LLaVA-Video & VAST & InternVideo2.5 & QwenVL2.5 & Reference \\
    \midrule
    % \rowcolor{gray!9} \multicolumn{5}{c}{\textit{VideoLLM}} \\
        \tiny
        The video begins with a view of a historic cathedral engulfed in flames, with thick smoke billowing into the sky. The cathedral's spire is prominently visible, and the surrounding buildings are also affected by the fire. The scene is dramatic, with the fire raging intensely and the smoke creating a dark cloud over the area. The video then transitions to a closer view of the cathedral's spire, which is now almost entirely consumed by the fire. The flames are bright and fierce, illuminating the intricate details of the spire. The smoke continues to rise, obscuring parts of the structure. The video progresses to show the spire collapsing under the heat, with the flames still burning fiercely. The final frames focus on the aftermath of the collapse, with the remains of the spire and the surrounding area still ablaze. The video concludes with a wider shot of the cathedral, showing the extent of the damage and the ongoing fire.
    & 
        \tiny fire is burning on a tall tower that has been burning 
    & 
        \tiny 
        The video captures the catastrophic fire engulfing the Notre Dame Cathedral in Paris. The iconic structure is ablaze, with flames and smoke billowing from its spire and roof. The fire's orange glow contrasts with the grey sky, while thick smoke rises into the air. The surrounding environment appears calm, with no visible people or firefighters in the immediate vicinity of the blaze. The cathedral's gothic architecture is partially visible, with its skeletal framework exposed by the fire. The scene is somber, highlighting the loss of a historical landmark.
    & 
        \tiny 
        The video depicts a catastrophic fire engulfing a historic cathedral, likely Notre-Dame in Paris, France. The flames are intense and widespread, with thick smoke billowing into the sky. The iconic spire of the cathedral is visibly on fire, with the structure glowing red from the heat. As the fire rages, the spire begins to weaken and eventually collapses, sending a cascade of debris downward. The surrounding buildings and trees are partially obscured by the smoke, emphasizing the scale and severity of the blaze. The scene is both dramatic and tragic, capturing a moment of significant destruction to a beloved landmark.
    & 
        \tiny 
        A structural fire broke out in the roof space of the Notre-Dame de Paris, a medieval Catholic cathedral in Paris, France. By the time the fire was extinguished, the cathedral's wooden spire, called a fléche, collapsed, most of the wooden roof had been destroyed, and the cathedral's upper walls were severely damaged. Extensive damage to the cathedral's interior was prevented by the vaulted stone ceiling, which largely contained the burning roof as it collapsed. However, some works of art and religious relics suffered smoke damage and some exterior art was damaged or destroyed. The cathedral's altar and three 13th-century rose windows suffered little or no damage. There were emergency workers present during the fire. The fire contaminated the site and nearby areas of Paris with toxic dust. French president Emmanuel Macron set a deadline for the restoration of the cathedral.
    \\
    \bottomrule
    \end{tabular}
    \caption{Qualitative Results for VideoLLMs on the Notre-Dame Fire query. The results of the table are from video ID: 1117864206475218944. \textbf{Bold text represents model hallucinations.} This is any information included in the article that is not in any of the videos.}
    \label{tab:notre-dame-videollm}
\end{table*}

\begin{table*}[]
    \centering
    \begin{tabular}{|p{0.16\linewidth}|p{0.16\linewidth}|p{0.16\linewidth}|p{0.16\linewidth}|p{0.16\linewidth}|}
         LLaVA-Video & VAST & InternVideo2.5 & QwenVL2.5 & Reference \\
    \midrule
    % \rowcolor{gray!9} \multicolumn{5}{c}{\textit{Article Generation}} \\
        \tiny 
        On April 15, 2019, a devastating fire broke out at Notre-Dame Cathedral in Paris, one of the world's most famous and historic Gothic landmarks. The blaze, which began in the attic, rapidly spread, causing significant damage to the cathedral's wooden frame and leading to the collapse of its spire. The fire raged for several hours, with thick plumes of smoke and flames visible from across the city. Despite the extensive damage, firefighters managed to save the cathedral's main structure, including the two bell towers. \textbf{The incident drew widespread international attention and prompted pledges of support for the reconstruction efforts from global leaders and private donors.} President Emmanuel Macron vowed that the cathedral would be restored, \textbf{and initial assessments indicated that the restoration process could take years to complete. The fire raised questions about the safety and maintenance of historical sites and the methods used to protect them from such disasters. }
    & 
        \tiny
        A major fire engulfed a prominent building in a French city, drawing a crowd of onlookers who gathered to witness the event. Flames and thick smoke were visible from various parts of the structure, including its roof and upper levels. The fire, which broke out in the middle of the building, \textbf{also affected a nearby church, causing significant damage.} Emergency services responded to the scene, while local residents and visitors stopped to observe and document the incident, some taking photographs and discussing the unfolding situation. \textbf{The cause of the fire and the extent of the damage remained under investigation.}
    & 
        \tiny 
        A catastrophic fire engulfed the Notre-Dame Cathedral in Paris \textbf{on April 15, 2019}, causing extensive damage to the historic structure. Flames and thick smoke billowed from the roof and spire, which partially collapsed during the blaze. \textbf{The fire, which began in the attic, spread rapidly, threatening the integrity of the entire building.} Despite the intense heat and the structural damage, the main facade and two bell towers remained standing. Emergency responders, including hundreds of firefighters, worked tirelessly throughout the night to control the fire. The event drew large crowds of onlookers and prompted a widespread outpouring of grief and solidarity, both locally and internationally. \textbf{The cause of the fire was initially suspected to be linked to renovation work being carried out on the cathedral.} The aftermath revealed a charred interior and significant structural damage, though some elements, such as the stained glass windows, survived. Restoration efforts were immediately pledged by the French government \textbf{and private donors.}
    & 
        \tiny 
        \textbf{On April 15, 2019}, a catastrophic fire broke out at Notre-Dame Cathedral in Paris, France, causing extensive damage to the historic landmark. The blaze, which erupted in the afternoon, rapidly spread through the cathedral's roof and spire, leading to the collapse of the iconic spire and significant portions of the roof. Thick plumes of smoke billowed into the sky, visible from across the city, as firefighters worked tirelessly to contain the inferno. The fire drew a large crowd of concerned onlookers, including Parisians and tourists, who gathered to watch the unfolding disaster, many expressing their shock and sorrow. Despite the extensive damage, the structural integrity of the cathedral's main walls and towers was preserved, and key religious artifacts, including the cross and some statues, remained intact. President Emmanuel Macron pledged to rebuild the cathedral, \textbf{which is a UNESCO World Heritage site and a symbol of French heritage and culture. The cause of the fire was initially suspected to be linked to ongoing renovation work, and investigations were launched to determine the exact cause. The event highlighted the global significance of Notre-Dame and the profound impact of its partial destruction on both the local community and the international community.}
    & 
        \tiny
        A structural fire broke out in the roof space of the Notre-Dame de Paris, a medieval Catholic cathedral in Paris, France. By the time the fire was extinguished, the cathedral's wooden spire, called a fléche, collapsed, most of the wooden roof had been destroyed, and the cathedral's upper walls were severely damaged. Extensive damage to the cathedral's interior was prevented by the vaulted stone ceiling, which largely contained the burning roof as it collapsed. However, some works of art and religious relics suffered smoke damage and some exterior art was damaged or destroyed. The cathedral's altar and three 13th-century rose windows suffered little or no damage. There were emergency workers present during the fire. The fire contaminated the site and nearby areas of Paris with toxic dust. French president Emmanuel Macron set a deadline for the restoration of the cathedral. \\
    \bottomrule
    \end{tabular}
    \caption{Qualitative Results for Article Generation on the Notre-Dame Fire query. \textbf{Bold text represents model hallucinations.} This is any information included in the article that is not in any of the videos.}
    \label{tab:notre-dame-ag}
\end{table*}

\begin{table*}[]
    \centering
    \begin{tabular}{|p{0.16\linewidth}|p{0.16\linewidth}|p{0.16\linewidth}|p{0.16\linewidth}|p{0.16\linewidth}|}
         LLaVA-Video & VAST & InternVideo2.5 & QwenVL2.5 & Reference \\
    \midrule
    % \rowcolor{gray!9} \multicolumn{5}{c}{\textit{\methodname}} \\
        \tiny
        \textbf{On April 15, 2019}, a massive fire broke out at Notre-Dame Cathedral in Paris, one of the most iconic and historically significant Gothic cathedrals \textbf{in the world}. The blaze, which \textbf{began in the attic} of the cathedral, rapidly spread, causing the collapse of the spire and significant damage to the roof and interior. Thick black smoke billowed into the sky, and the flames were visible from across the city. Despite the intensity of the fire, firefighters managed to control and eventually extinguish the blaze by early morning. The incident drew a large crowd of onlookers, many of whom gathered along the Seine River and near the cathedral, some singing hymns in response to the tragedy. President Emmanuel Macron pledged to rebuild the cathedral, \textbf{and restoration efforts were soon underway, with the goal of preserving and restoring this cherished landmark.} The fire caused extensive damage, including the loss of the spire and a significant portion of the roof, but the main structure and many valuable artifacts were saved. \textbf{The event highlighted the cathedral's enduring cultural and spiritual significance to both the French people and the global community.}
    &
        \tiny
        \methodname was not run on VAST due to poor extraction capabilities.

    & 
        \tiny
        A catastrophic fire engulfed the Notre Dame Cathedral in Paris \textbf{on April 15, 2019}, causing significant damage to the historic structure. Flames and thick smoke billowed from the roof and spire, which partially collapsed during the blaze. The fire, which spread rapidly, was visible from afar, drawing large crowds of onlookers who watched in shock and dismay. Emergency responders, including firefighters, were quickly deployed to the scene, working tirelessly to control the inferno and prevent further damage. Despite their efforts, the cathedral suffered extensive damage, with the roof and spire being among the most affected areas. The incident occurred during the evening, and the surrounding area was evacuated to ensure public safety. The aftermath revealed a charred and debris-filled interior, with the iconic spire and much of the roof destroyed. \textbf{The fire's cause was initially unknown, but investigations were launched to determine the origin and circumstances of the disaster.} The event sparked a global outpouring of support and grief, with many gathering to mourn the loss of this cultural and religious landmark.
    & 
        \tiny
        \textbf{On April 15, 2019,} a catastrophic fire broke out at Notre-Dame Cathedral in Paris, France, causing extensive damage to the historic landmark. The fire, \textbf{which began in the attic} of the cathedral, rapidly spread, engulfing the roof and the iconic spire, which collapsed in a dramatic fall witnessed by a shocked crowd gathered outside. The cathedral, a masterpiece of Gothic architecture \textbf{dating back to the 12th century}, suffered significant structural damage, with the roof and spire completely destroyed. Despite the intense blaze, the main stone structure, including the two bell towers, remained standing. Firefighters worked tirelessly throughout the night to contain the fire, \textbf{using advanced equipment and techniques to prevent further damage}. \textbf{The event drew global attention, with people around the world expressing sorrow and solidarity}. In the aftermath, French President Emmanuel Macron pledged to rebuild the cathedral, \textbf{and a fundraising campaign was launched to support the restoration efforts.} \textbf{The fire raised questions about the safety and preservation of historic landmarks and highlighted the cultural and spiritual significance of Notre-Dame to both France and the international community.}

    & 
        \tiny
        A structural fire broke out in the roof space of the Notre-Dame de Paris, a medieval Catholic cathedral in Paris, France. By the time the fire was extinguished, the cathedral's wooden spire, called a fléche, collapsed, most of the wooden roof had been destroyed, and the cathedral's upper walls were severely damaged. Extensive damage to the cathedral's interior was prevented by the vaulted stone ceiling, which largely contained the burning roof as it collapsed. However, some works of art and religious relics suffered smoke damage and some exterior art was damaged or destroyed. The cathedral's altar and three 13th-century rose windows suffered little or no damage. There were emergency workers present during the fire. The fire contaminated the site and nearby areas of Paris with toxic dust. French president Emmanuel Macron set a deadline for the restoration of the cathedral. \\
    \bottomrule
    \end{tabular}
    \caption{Qualitative Results for \methodname on the Notre-Dame Fire query. \textbf{Bold text represents model hallucinations.} This is any information included in the article that is not in any of the videos.}
    \label{tab:notre-dame-cag}
\end{table*}
\paragraph{Qualitative Results for Model Variations}
In this section we present additional qualitative examples from the different generation methods. We present these for the Notre-Dame fire query against the reference article (\autoref{tab:notre-dame-videollm}, \autoref{tab:notre-dame-ag}, \autoref{tab:notre-dame-cag}). 

\paragraph{Qualitative Results for Audio vs. Video}
We also present the qualitative differences between video only and video+audio CAG results in \autoref{tab:video_v_audio1}, \autoref{tab:video_v_audio2}, and \autoref{tab:video_v_audio3}. In these tables you can see that the articles produced using transcripts are shorter than the articles based only on video content. Numerically, the average length of a video-only article is 206.36 tokens and the average length of a article with transcript provided is 163.90 tokens. 

\begin{table*}[]
    \centering
    \begin{tabular}{p{0.45\linewidth}|p{0.45\linewidth}}
    \toprule
         Video & Audio \\
    \midrule
        \small
        On April 15, 2019, a catastrophic fire broke out at Notre-Dame Cathedral in Paris, France, causing extensive damage to the historic landmark. The fire, which began in the attic of the cathedral, rapidly spread, engulfing the roof and the iconic spire, which collapsed in a dramatic fall witnessed by a shocked crowd gathered outside. The cathedral, a masterpiece of Gothic architecture dating back to the 12th century, suffered significant structural damage, with the roof and spire completely destroyed. Despite the intense blaze, the main stone structure, including the two bell towers, remained standing. Firefighters worked tirelessly throughout the night to contain the fire, using advanced equipment and techniques to prevent further damage. The event drew global attention, with people around the world expressing sorrow and solidarity. In the aftermath, French President Emmanuel Macron pledged to rebuild the cathedral, and a fundraising campaign was launched to support the restoration efforts. The fire raised questions about the safety and preservation of historic landmarks and highlighted the cultural and spiritual significance of Notre-Dame to both France and the international community.
    & 
        \small
        On April 15, 2019, a catastrophic fire broke out at Notre-Dame Cathedral in Paris, France, causing significant damage to the historic landmark. The blaze, which started in the attic, rapidly spread, engulfing the roof and the iconic spire, which subsequently collapsed. Thick plumes of smoke and intense flames were visible from afar, drawing crowds of onlookers who watched in distress and solidarity. The fire, which lasted for several hours, was eventually brought under control by hundreds of firefighters. Despite the extensive damage, the main structure of the cathedral, including the two bell towers, survived. President Emmanuel Macron pledged to rebuild the cathedral, and international support for the restoration efforts was swift and substantial. The incident highlighted the cultural and historical significance of Notre-Dame, one of the world's most recognized and cherished Gothic cathedrals.
        
    \\
    
    \bottomrule
    \end{tabular}
    \caption{QwenVL+\methodname with and without audio. Notre-Dame Fire}
    \label{tab:video_v_audio1}
\end{table*}

\begin{table*}[]
    \centering
    \begin{tabular}{p{0.45\linewidth}|p{0.45\linewidth}}
    \toprule
         Video & Audio \\
    \midrule
        \small
        The 2022 Georgia Senate election was a closely contested race between incumbent Democrat Raphael Warnock and Republican challenger Herschel Walker. After the initial election on November 8, 2022, neither candidate secured the required 50\% majority, leading to a runoff election on December 6, 2022. In the initial vote, Warnock received 49.2\% (1,944,893 votes) and Walker received 48.7\% (1,927,402 votes), with a difference of 17,491 votes. An independent candidate, Chase Oliver, received 2.1\% of the vote. The runoff election, which determined the ultimate winner, saw a significant turnout, with over 2 million early votes cast, highlighting the high level of voter engagement. The outcome of the runoff was crucial for the balance of power in the U.S. Senate, as it would determine whether the Democrats maintained their slim majority or if the Republicans gained control. Ultimately, Raphael Warnock was declared the winner, securing his re-election to the Senate.
    & 
        \small
        The 2022 Georgia Senate election was a closely contested race between incumbent Democratic Senator Raphael Warnock and Republican challenger Herschel Walker. With neither candidate reaching the 50\% threshold required to avoid a runoff, the election will proceed to a runoff on December 6, 2022. Preliminary results showed Warnock leading with 49.2\% of the vote (1,944,893 votes) and Walker trailing slightly with 48.7\% (1,927,402 votes). Independent candidate Chase Oliver received 2.1\% of the vote (81,130 votes). The outcome of the runoff is expected to have significant implications for control of the U.S. Senate and for voting rights legislation, as well as for the broader political landscape in Georgia and nationally.
        
    \\
    
    \bottomrule
    \end{tabular}
    \caption{QwenVL+\methodname with and without audio. Georgia Senate Election}
    \label{tab:video_v_audio2}
\end{table*}

\begin{table*}[]
    \centering
    \begin{tabular}{p{0.45\linewidth}|p{0.45\linewidth}}
    \toprule
         Video & Audio \\
    \midrule
        \small
        Hurricane Irma was a powerful and destructive Category 5 hurricane that caused catastrophic damage across the Caribbean and parts of the southeastern United States in September 2017. Originating as a tropical wave, Irma rapidly intensified into a Category 5 storm with maximum sustained winds of 180 miles per hour, becoming the most powerful hurricane ever recorded in the open Atlantic. The storm made landfall in the Caribbean, particularly devastating the islands of St. Martin, the British Virgin Islands, and Barbuda. It then moved northwest, causing extensive damage in the Bahamas and Cuba before making landfall in Southwest Florida on September 10. The hurricane brought torrential rain, storm surges, and high winds, resulting in widespread flooding, structural damage, and power outages. In Florida, cities such as Miami Beach and Tampa experienced severe flooding and wind damage. The storm's impact was felt across multiple states, leading to widespread evacuations and a significant humanitarian response. Recovery efforts were complicated by the extensive damage to infrastructure, including roads, bridges, and power lines. The hurricane's aftermath highlighted the resilience of affected communities and the critical need for coordinated disaster preparedness and response.
    & 
        \small
        Hurricane Irma was a powerful and destructive Category 5 hurricane that caused widespread damage across the Caribbean and the southeastern United States in September 2017. Originating as a tropical wave, Irma rapidly intensified into a Category 5 storm with maximum sustained winds reaching 180 miles per hour. The hurricane made landfall in the Leeward Islands, causing catastrophic damage, particularly on the islands of Barbuda and St. Martin. It then moved through the Turks and Caicos, the Bahamas, and eventually struck the Florida Keys and the southwestern coast of Florida. The storm brought torrential rainfall, storm surges, and strong winds, leading to extensive flooding and structural damage. In the aftermath, communities faced significant challenges, including the loss of electricity, water, and communication services. Recovery efforts were hampered by the scale of the destruction, and many areas required substantial aid and support to rebuild.
        
    \\
    
    \bottomrule
    \end{tabular}
    \caption{QwenVL+\methodname with and without audio. Hurricane Irma}
    \label{tab:video_v_audio3}
\end{table*}

\paragraph{Qualitative Results for 2025 Subset}
We also present the qualitative differences between CAG and LLM responses for the 2025 subset of \dsetname. The LLM is prompted with "Generate a Wikipedia article about [event]." \autoref{tab:2025-qual} shows the comparison in these results. However, the LLM results are poor, filled with ungrounded and not factual information. In the main text, it's noted that the Rouge and BertScores show similar performance of the LLM to \cagr. However, these qualitative examples show that the LLM is only matching the style of Wikipedia articles and not generating meaningful and factual text.

\begin{table*}[]
    \centering
    \begin{tabular}{p{0.1\linewidth}|p{0.4\linewidth}|p{0.4\linewidth}}
    \toprule
         \textbf{Topic} & \textbf{LLM} & \textbf{\methodname} \\
    \midrule
    Liberation Day Tariffs &
    \tiny 
        Liberation Day Tariffs, also known as Vrijheidsdag Tariffs, are special rates and discounts offered by various businesses and organizations in the Netherlands on May 5th, Liberation Day (Vrijheidsdag). This day commemorates the end of the Nazi occupation during World War II in 1945. The tariffs typically include reduced prices on public transport, free admission to museums, and special events and activities designed to encourage participation and celebration among the Dutch population. These offers aim to make the commemoration accessible and inclusive, fostering a sense of community and national pride.
    & 
    \tiny 
        On April 2, 2025, President Donald Trump delivered a speech in Washington, D.C., announcing new tariffs on imports from several countries, including China, the European Union, Vietnam, Taiwan, Japan, India, South Korea, Thailand, Switzerland, Indonesia, and Malaysia. The tariffs, ranging from 20\% to 36\%, were designed to address what the administration perceived as unfair trade practices and to encourage domestic manufacturing. Trump emphasized that the U.S. would not impose a full 100\% reciprocal tariff, stating his intention to be \"kind to the world\" while protecting American industries. The announcement, made after the stock market closing bell, had already contributed to market volatility, with significant losses reported in the stock market, including a \$3 trillion drop in value, referred to as a \"stock market armageddon.\" Financial analysts, such as those from JPMorgan Chase \& Co., raised the recession risk forecast to 60\%, up from 40\% prior to the announcement. The legal and economic implications of these tariffs were widely discussed, with experts debating the constitutionality of the President's actions under the International Emergency Economic Powers Act (IEEPA) and the potential long-term impacts on global trade dynamics and consumer prices.
    \\ 
    \midrule
    Blue Ghost Mission 1 & 
    \tiny
        The Blue Ghost Mission 1 was the inaugural flight of the Blue Ghost spacecraft, developed by Astra, an American aerospace manufacturer. Launched from the Mid-Atlantic Regional Spaceport (MARS) on Wallops Island, Virginia, the mission aimed to demonstrate the capabilities of the reusable suborbital vehicle designed for research and technology development. The spacecraft, which can carry payloads up to 150 kilograms, successfully reached its intended altitude and completed a series of tests before safely returning to Earth. The mission marked a significant milestone in Astra's efforts to provide cost-effective and flexible access to space for scientific and commercial applications.
    &
    \tiny
        The Blue Ghost M1 lunar lander, developed by Firefly Aerospace, successfully landed on the Moon on February 24, 2025, as part of NASA's Artemis program. The mission, which aimed to advance lunar exploration and scientific research, was a significant milestone in space exploration. The spacecraft, equipped with advanced scientific instruments, was designed and assembled at Firefly Aerospace's industrial facility, where engineers and technicians worked meticulously to ensure its precision and reliability. The landing process was closely monitored from the company's control room, where a team of experts tracked the lander's descent using detailed telemetry data. The successful touchdown was met with jubilation among the team members, marking a testament to their dedication and innovation in space technology. The mission not only contributed to the ongoing exploration of the Moon but also highlighted the capabilities of private aerospace companies in collaboration with government space agencies.
    \\
    \midrule
    Canadian Federal Election & 
    \tiny 
        The 2025 Canadian federal election was held on October 21, 2025, to elect members of the House of Commons of Canada, the lower house of the bicameral Parliament of Canada. This election determined the composition of the 45th Canadian Parliament and the next Prime Minister of Canada. The Liberal Party, led by incumbent Prime Minister Justin Trudeau, faced off against the Conservative Party, led by Pierre Poilievre, among other smaller parties. Key issues included economic management, healthcare reform, and climate change policies. The election saw high voter turnout and significant debate over the direction of national policy, reflecting deep divisions within the electorate. Results from the election were closely watched, with implications for both domestic and international affairs.
    & 
    \tiny 
        The 2025 Canadian federal election, held on April 29, 2025, resulted in a fragmented Parliament with no single party achieving a majority. The Liberal Party, led by the incumbent Prime Minister, secured 119 seats, while the Conservative Party, under the leadership of Pierre Poilievre, gained over 20 seats, bringing their total to 83 and receiving the highest share of votes since 1988. The Bloc Qu\u00e9b\u00e9cois won 15 seats, the New Democratic Party (NDP) secured only 4 seats, and the Green Party failed to win any seats. The election saw significant shifts in voter support, particularly from the NDP to the Liberals, and created a challenging political environment with neither the NDP nor the Liberals able to form a coalition government. Pierre Poilievre, in his post-election speech, acknowledged the party's progress but emphasized the need for continued effort and change, looking ahead to future elections. The overall atmosphere during election night was one of excitement and celebration, especially among Liberal supporters, as they awaited the final results.
    \\
    \bottomrule
    \end{tabular}
    \caption{Qualitative outputs on a sample of the 2025 subset of WikiVideo for the LLM and \methodname outputs. 
    % \textbf{Bold} text represents information not in the videos.
    }
    \label{tab:2025-qual}
\end{table*}

\section{Incorporating Audio}
\label{append:modality_experiment} 
\begin{table}
    \centering
    \begin{tabular}{lc|cccc}
    \toprule
         \textbf{Method} & \textbf{VLM} & \textbf{R1} & \textbf{BS} & \textbf{Arg} & \textbf{AS} \\
    \midrule
        % LLaVA-Video-7B & \xmark & \cmark & - & - & - & - & - & - \\  
        \multirow{1}{*}{CAT-0} 
        % & LV & 5.2 & 79.5 & 20.3 & 6.0 \\
        % % Instruct-V2Xum & \xmark & \cmark & - & - & - & - & - & - \\
        % & IV & 11.6 & 80.7 & 18.7 & 8.5 \\
        & QVL & 11.1 & 81.6 & 22.1 & 7.50 \\
        % LLaVA-Video-7B & \cmark & \cmark & - & - & - & - & - & - \\
        \multirow{1}{*}{\cagr}
        % & LV & 29.4 & 77.3 & 25.6 & \underline{13.2} \\
        % % VAST & \cmark & \cmark &  - & - & - & - & - & - \\
        % % Instruct-V2Xum & \cmark & \cmark & - & - & - & - & - & - \\
        %  & IV & \underline{32.8} & \underline{85.8} & 24.8 & 13.1 \\
        & QVL & 32.1 & 85.7 & \underline{26.3} & 12.6 \\
    \midrule
    \rowcolor{gray!15}
        \cagr & QVL & \textbf{34.0} & \textbf{86.4} & \textbf{30.8} & \textbf{14.3} \\
    \bottomrule
    % \rowcolor{gray!25}
    %     Human & & & 40.05 & 13.33 & 20.79 & 86.61 \\
    % \rowcolor{gray!25}
    %     Wikipedia & & & 64.53 & 48.27 & 53.25 & 90.47 \\
    \end{tabular}
    \caption{\dsetname article generation results for \methodname  \emph{with} audio inputs. Bottom row shows \methodname \emph{without} audio (copied from \autoref{tab:main_result}), which performs best.}
    \label{tab:modality_results}
    % \vspace{-0.5em}
\end{table}

Articles in \dsetname have many claims grounded partly or only in videos' audio signal~(\autoref{tab:data_stats}). Here, we consider the impact of adding audio information as additional input to \methodname and to \concatg. For all methods except for VAST--which takes raw audio input--we transcribe the audio of each video using whisper-v3 large \citep{radford2022whisper}. For the \concatg baseline, we provide the transcription as additional input alongside the instruction and frames to the VideoLLM. For \methodname, we provide the transcriptions for each video together with the per-video summaries as input to the text-only aggregator LLM.

\autoref{tab:modality_results} presents the results. Similar to the previous experiment, we find that \concatg continues to yield poor quality articles, even with audio information. Notably, however, both \concatg and \methodname consistently obtain worse results with audio inputs than without (\autoref{tab:main_result})---despite the sizable fraction of audio-support claims in \dsetname. For \concatg, this may partly be explained by the fact that the pretraining data for the VideoLLMs we study does not include audio transcripts, and thus the prompts that incorporate them are out-of-distribution. For \methodname, we found that including audio transcripts tended to result in substantially shorter final articles (avg.\ $\sim 164$ tokens) than omitting them (avg.\ $\sim 206$ tokens)---suggesting that the former may be less thorough in their coverage of the event relative to the references, leading to lower metric scores (\autoref{append:outputs} has quantitative examples). Identifying ways to more effectively incorporate audio into the \dsetname task thus constitutes an intriguing direction for future work.

\section{Additional Results}
\label{append:numbers}

We report more statistics in \autoref{tab:all_numbers} to show the varying performance across model sizes and variations. We report LLaVA-Video-7B,72B and QwenVL2.5-3B,7B,72B as well as Qwen2.5-32B,72B for article synthesis.

\begin{table*}[t]
    \centering
    \begin{tabular}{cc|cccccc}
    \toprule
        Method & VideoLLM & R1 & R2 & RL & BS & Arg & AS \\
    \midrule
    % \multirow{7}{*}{\concatg}
        % & LLaVA-Video-7B & \xmark & \xmark & - & - & - & 79.35 & - & - \\
        & LLaVA-Video-7B & 4.24 & 1.02 & 2.93 & 79.35 & 20.07 & 6.26\\
        & LLaVA-Video-72B & 7.34 & 1.60 & 4.78 & 71.99 & 19.31 & 5.08 \\
        & VAST & 16.62 & 1.71 & 11.19 & 80.55 & 8.04 & 7.13 \\
        \concatg & InternVideo2.5 & 11.85 & 2.32 & 7.90 & 80.78 & 18.33 & 9.53 \\
        & QwenVL2.5-3B & 9.60 & 2.36 & 6.27 & 80.80 & 20.22 & 7.73\\
        & QwenVL2.5-7B & 9.82 & 2.62 & 6.25 & 81.27 & 21.28 & 9.04\\
        & QwenVL2.5-72B & 11.34 & 3.13 & 7.06 & 81.60 & 23.72 & 8.01 \\
    \midrule
    % \multirow{3}{*}{\concatr}
        & LLaVA-Video & 6.36 & 1.51 & 4.22 & 80.03 & 21.34 & 5.50\\
        % & VAST & \\
        % Instruct-V2Xum & \xmark & \xmark & - & - & - & - & - & - \\
        \concatr & InternVideo2.5 & 6.93 & 1.68 & 4.83 & 79.62 & 22.48 & 6.19 \\
        & QwenVL2.5 & 8.38 & 2.71 & 5.49 & 81.94 & 22.89 & 7.17\\
    \midrule
    % \multirow{7}{*}{\summg+32B}
        & LLaVA-Video-7B & 31.04 & 7.96 & 18.14 & 85.65 & 24.49 & 18.71\\
        & LLaVA-Video-72B & 28.55 & 7.10 & 16.00 & 77.00 & 22.33 & 15.54 \\
        & VAST & 16.80 & 1.12 & 11.18 & 81.63 & 9.17 & 14.01 \\
        \summg+32B & InternVideo2.5 & 28.11 & 6.06 & 16.94 & 84.80 & 22.21 & 16.59\\
        & QwenVL2.5-3B & 30.78 & 7.87 & 18.19 & 85.39 & 24.32 & 16.37\\
        & QwenVL2.5-7B & 31.64 & 7.88 & 18.13 & 85.59 & 24.35 & 16.31 \\
        & QwenVL2.5-72B & 32.59 & 8.86 & 18.89 & 85.81 & 26.60 & 15.71 \\
    \midrule
    % \multirow{7}{*}{\summg+72B}
        & LLaVA-Video-7B & 34.87 & 10.72 & 20.18 & 86.44 & 28.09 & 16.34\\
        & LLaVA-Video-72B & 30.02 & 8.68 & 17.59 & 77.59 & 26.21 & 13.51 \\
        & VAST & 19.55 & 1.45 & 12.40 & 82.21 & 11.23 & 10.87 \\
        \summg+72B & InternVideo2.5 & 32.54 & 8.98 & 19.47 & 85.82 & 25.65 & 17.58 \\
        & QwenVL2.5-3B & 32.92 & 10.00 & 19.37 & 85.95 & 27.44 & 16.27 \\
        & QwenVL2.5-7B & 34.01 & 10.05 & 19.47 & 86.24 & 26.97 & 16.72 \\
        & QwenVL2.5-72B & 33.58 & 10.15 & 19.15 & 86.18 & 28.97 &  15.63\\
    \midrule
    % \multirow{3}{*}{\methodname}
        & LLaVA-Video & 33.38 & 10.05 & 19.44 & 84.55 & 28.26 & 15.23 \\
        % & VAST &  - & - & - & - & - & - \\
        \methodname & InternVideo2.5 & 33.91 & 9.58 & 20.07 & 86.13 & 27.01 & 14.23 \\
        & QwenVL 2.5 & 33.96 & 10.90 & 19.45 & 86.35 & 30.77 & 14.29 \\
    \bottomrule
    \end{tabular}
    \caption{Vision only results by method.RePrompt results are \textbf{only} the outputs from the follow-up questions and does not include the generic captions.}
    \label{tab:all_numbers}
\end{table*}

\paragraph{Per-event type results}
\begin{table*}[]
    \centering
    \addtolength{\tabcolsep}{-0.2em}
    \begin{tabular}{c|cc|cc|cc|cc|cc|cc|cc|cc}
    \toprule
         & \multicolumn{2}{c}{All}& \multicolumn{2}{c}{Sport} & \multicolumn{2}{c}{Disaster} & \multicolumn{2}{c}{Election} & \multicolumn{2}{c}{Social} & \multicolumn{2}{c}{Demonst} & \multicolumn{2}{c}{Discover} & \multicolumn{2}{c}{Political}\\
        Model & ED & EX & ED & EX & ED & EX & ED & EX & ED & EX & ED & EX & ED & EX & ED & EX \\
    \midrule
        LV & 28 & 09 & 35 & 11 & 24 & 04 & 58 & 44 & 36 & 08 & 17 & 00 & 34 & 13 & 26 & 08\\
        
        IV & 27 & 06 & 40 & 17 & 24 & 03 & 49 & 25 & 42 & 00 & 13 & 00 & 30 & 09 & 23 & 04\\
        
        QVL & 31 & 11 & 50 & 32 & 29 & 06 & 62 & 50 & 38 & 08 & 18 & 00 & 31 & 10 & 25 & 05\\
    \bottomrule
    \end{tabular}
    \caption{Argument F1 by MultiVENT-G Event Type for \methodname Sport, Disaster, Election
    Social: Social Event, Demonst: Demonstration, Discover: Discovery OR Launch, Political: Political Development. ED: Edit Distance, EX: Exact Match}
    \label{tab:extraction_scores}
\end{table*}
In \autoref{tab:extraction_scores}, we breakdown the argument F1 scores for each VideoLLM+\methodname combination. Here we see the highest F1 scores in the most commonly recognizable events: elections and sports. These events are often professionally broadcast and the entities that participate in these events are ``high-resource'' visual concepts. However, in events like disasters and demonstrations, we see a decrease in F1, especially in exact match as there are no longer high-resource entities to identify or heavily populated OCR content. 

\section{Human Analysis}
\label{append:human_analysis}

To provide an upper-bound on model performance, we recruit 3 fluent english speakers to write 3 articles. These annotators receive the information request and the set of ``oracle'' relevant videos and are instructed to write the article from this information. This human annotation is fundamentally different than our data collection process because instead of grounding and `discriminating' against an existing text, the annotators perform the same task as \methodname taking the videos and writing information from them. To create the human generated articles, we provide annotators the relevant videos and instruct them to write the lead of a Wikipedia article. An interesting note from this experiment is we notice the annotators perform article writing similar to \methodname, taking notes on each video before aggregating them in an article. 

In \autoref{tab:human_eval}, we baseline human performance against the original Wikipedia article as the predicted article and the best method (\methodname+QwenVL). We observe that the current metrics for the task don't accurately capture the quality of the human written article, which has no hallucinations and is fully follows the constraint of only including video content. We show these results qualitatively in \autoref{tab:model_human_ref}, \autoref{tab:model_human_ref1}, and \autoref{tab:model_human_ref2}.
% \begin{table}[]
%     \centering
%     \begin{tabular}{cc|cccccc}
%         Method & Model & R1 & R2 & RL & BS & CR & CS \\
%     \midrule
%     %     Evolution & Human & 40.05 & 13.33 & 20.79 & 86.61 \\
%     % - & - & 38.54 & 12.91 & 20.11 & 86.34 \\
%     % \rowcolor{gray!25}
%     % Wikipedia & Human & 64.53 & 48.27 & 53.25 & 90.47 \\ & 
%     \end{tabular}
%     \caption{Caption}
%     \label{tab:human_eval}
% \end{table}

% \begin{wraptable}{r}{8.5cm}
% \centering
% % \vspace{-2.5em}
% \begin{tabular}{cc|cccc}
% \toprule  
%     Method & Model & R1 & BS & Arg & AS \\
% \midrule
%     \methodname & QwenVL & 40.57 & 86.77\\
%     {\small RAG+\methodname} & QwenVL & 23.84 & 77.85 \\
% \midrule 
%     Evolution & Human &  38.54 & 86.34 \\
%     Wikipedia & Human & 64.53 & 90.47 \\
    
% \bottomrule
% \end{tabular}
% \caption{Comparison to human performance.}
% \label{tab:human_eval}
% % \vspace{-1em}
% \end{wraptable} 

\begin{table*}[]
    \centering
    \begin{tabular}{cc|cccc}
    \toprule  
        Method & VideoLLM & R1 & BS & Arg & AS \\
    \midrule
        \methodname & QwenVL & 40.57 & 86.77 & 30.80 & 14.29\\
        {\small RAG+\methodname} & QwenVL & 23.84 & 77.85 & 20.65 & \phantom{0}9.01 \\
    \midrule 
        \multicolumn{2}{c}{Human Annotator} &  38.54 & 86.34 & 39.07 & 17.17\\
        \multicolumn{2}{c}{Wikipedia} & 64.53 & 90.47 & 62.82 & 23.53 \\
        
    \bottomrule
    \end{tabular}
    \caption{Comparison to human performance.}
    \label{tab:human_eval}
\end{table*}

\begin{table*}[]
    \centering
    \begin{tabular}{p{0.3\linewidth}|p{0.3\linewidth}|p{0.3\linewidth}}
    \toprule
         \methodname & Human & Reference \\
    \midrule
        \tiny
        The 2022 Georgia Senate election was a closely contested race between incumbent Democrat Raphael Warnock and Republican challenger Herschel Walker. After the initial election on November 8, 2022, neither candidate secured the required 50\% majority, leading to a runoff election on December 6, 2022. In the initial vote, Warnock received 49.2\% (1,944,893 votes) and Walker received 48.7\% (1,927,402 votes), with a difference of 17,491 votes. An independent candidate, Chase Oliver, received 2.1\% of the vote. The runoff election, which determined the ultimate winner, saw a significant turnout, with over 2 million early votes cast, highlighting the high level of voter engagement. The outcome of the runoff was crucial for the balance of power in the U.S. Senate, as it would determine whether the Democrats maintained their slim majority or if the Republicans gained control. Ultimately, Raphael Warnock was declared the winner, securing his re-election to the Senate.
    &
        \tiny

        On December 6th, 2022, ABC news declared Raphael Warnock the winner of the Georgia Senate Race. Warnock, the incumbent Democrat, defeated his opponent, republican Hershel Walker, in a runoff election by a little more than 1\% of the vote. Warnock had initially won election in 2020, when he, along with President Joe Biden and Senator John Ossoff, led a surprise sweep of the presidential and senate elections in the historically conservative state. A month prior to the runoff, Warnock also held a slight lead over Walker on election day. At that point, with 99\% of the votes counted, Warnock led by 35,429 votes, 49.4\% to 48.5\%. Despite this, in Georgia a runoff is triggered if no candidate wins at least 50\% of the vote in the general election. With Libertarian candidate Chase Oliver pulling roughly 2\% of the vote, MSNBC and CNN both reported that neither Warnock nor Walker were able to reach this threshold in November. This result was generally expected, as polling averages from the weekend prior to the initial general election showed an extremely tight race, from anywhere between a 0.8\% margin from MSNBC to just a 0.1\% margin from FiveThirtyEight.
        Many battleground senate races were decided relatively early on Election day, with North Carolina, Ohio, and Florida all being called for Republicans, while Pennsylvania and New Hampshire were called for Democrats. However, after these initial results, both Georgia and Wisconsin, where Ron Johnson held a slight lead over Mandela Barnes, were too close to call, while Nevada and Arizona also took longer for a victor to be declared. Interestingly, in contrast to the Georgia senate race, the Georgia gubernatorial election was decided without a runoff, with Republican Brian Kemp winning re-election in his rematch with Democrat Stacey Abrams, indicating a large number of split-ticket voters. This came despite Abrams’ long-term investments into an activist-driven campaign. 

        In the lead up to the runoff election, Walker heavily emphasized Joe Biden’s low approval ratings and the economy, while Warnock focused on Walker’s lack of knowledge about issues and allegations of violence. The incumbent senator’s strategy focused on turning out the democratic base in cities, while the challenger focused on more conservative rural areas; both campaigns targeted moderate educated voters in the suburbs around Atlanta. In addition, Georgia Democrats won a key court battle to allow for early voting in runoff elections. This decision challenged SB 202, a 2021 state law that limited early voting periods and restricted weekend voting.

    & 
        \tiny
        The 2022 United States senate election in Georgia was held on November 8, 2022 to elect a member of the U.S. senate to represent the state of Georgia. The election was part of the 2022 midterm elections and featured incumbent democratic senator Raphael Warnock and Republican former football player Herschel Walker. Warnock, who had one a shortened seat in a special election held in 2020-21 defeated Walker in a runoff election and was re-elected for his first full term in office.
        
        Under Georgia's two-round election system, a runoff election occurs if no candidate receives more than 50\% of the vote. In the November 8 election, Warnock received 49.4\% of the vote and Walker received 48.5\%, triggering the December 6th runoff election. 
        
        Warnock's victory in the runoff was a state-wide victory for Democrats. Warnock's win gave Democrats a net gain of one seat in the 2022 midterms. The election was also one of many very close Senate elections during the 2022 midterms, with both Nevada and Wisconsin also having tight races.
        
    \\
    
    \bottomrule
    \end{tabular}
    \caption{Georgia Senate Election}
    \label{tab:model_human_ref}
\end{table*}

\begin{table*}[]
    \centering
    \begin{tabular}{p{0.3\linewidth}|p{0.3\linewidth}|p{0.3\linewidth}}
    \toprule
         \methodname & Human & Reference \\
    \midrule
        \tiny
        A significant earthquake with a magnitude of 7.0 struck Anchorage, Alaska, causing widespread disruption and damage. The National Weather Service confirmed the event and reassured the public that there was no tsunami threat. Footage from the aftermath showed damaged roads, disrupted infrastructure, and scenes of chaos, including a supermarket where items had fallen off shelves. The earthquake prompted school evacuations and the temporary closure of the Anchorage airport, as the air traffic control tower was evacuated. While the full extent of the damage remains unclear, ongoing assessments and recovery efforts are underway to address the impact of the quake.
    &
        \tiny
        A magnitude 7.0 earthquake hit southcentral Alaska on Friday, November 30, 2018. The earthquake hit approximately 7 miles north of Anchorage at 8:29 AM. Anchorage is Alaska’s largest city, with half of the state’s population living in the region. Following the earthquake, the National Tsunami Warning Center issued and then canceled a tsunami warning for the coastal zones of southern Alaska. Governor Bill Walker issued a disaster declaration. In the three hours after the earthquake, the US Geological Survey recorded at least 30 aftershocks with magnitudes ranging from 2.7 to 5.7; hundreds of aftershocks were eventually recorded. During the earthquake and its aftershocks, students and office workers sheltered in place under their desks, and some buildings were evacuated, including the air traffic control tower at the Anchorage airport. The earthquake caused major infrastructure damage across the city, according to the Anchorage police department. Impacts include damage to roadways and water mains, visible cracks in buildings, stores and homes in disarray, and one house fire caused by a damaged gas pipe. No deaths have been reported.

    & 
        \tiny
        On November 30, 2018 at 8:29 a.m. AKST (17:29 UTC), a magnitude 7.1 earthquake hit Anchorage in South Central Alaska. The earthquake's epicenter was 10 miles north of Anchorage and occurred at a depth of 29 miles. It was followed by a magnitude 5.7 aftershock. The National Tsunami Warning Center issued tsunami warnings for nearby costal areas, including Cook Inlet. The warnings were lifted shortly after being issued. 
        
    \\
    
    \bottomrule
    \end{tabular}
    \caption{Anchorage Earthquake}
    \label{tab:model_human_ref1}
\end{table*}

\begin{table*}[]
    \centering
    \begin{tabular}{p{0.3\linewidth}|p{0.3\linewidth}|p{0.3\linewidth}}
    \toprule
         \methodname & Human & Reference \\
    \midrule
        \tiny
        Hurricane Irma was a powerful and destructive Category 5 hurricane that caused widespread damage across the Caribbean and the southeastern United States in September 2017. Originating as a tropical wave, Irma rapidly intensified into a Category 5 storm with maximum sustained winds reaching 180 miles per hour. The hurricane made landfall in the Leeward Islands, causing catastrophic damage, particularly on the islands of Barbuda and St. Martin. It then moved through the Turks and Caicos, the Bahamas, and eventually struck the Florida Keys and the southwestern coast of Florida. The storm brought torrential rainfall, storm surges, and strong winds, leading to extensive flooding and structural damage. In the aftermath, communities faced significant challenges, including the loss of electricity, water, and communication services. Recovery efforts were hampered by the scale of the destruction, and many areas required substantial aid and support to rebuild.
    &
        \tiny
        Hurricane Irma was a record-setting Hurricane that struck the south Atlantic region in early September, 2017, and that caused widespread damage. At the time, Irma was the most powerful Hurricane recorded in the open Atlantic region and was the second-strongest ever to hit Cuba. Beginning as a tropic wave, Irma struck many places—including the Lesser Antilles, Dominica, Guadeloupe, Barbuda, Antigua, St. Martin, Puerto Rico, Haiti, the Dominican Republic, and Cuba—as a Category 5 storm—as well as Florida as a Category 3 storm. At its peak intensity, Irma reached more than 180 mph winds and caused severe flooding and damage to buildings and infrastructure throughout the region.

    & 
        \tiny
        Hurricane Irma was an extremely powerful Cape Verde hurricane that occurred in early September 2017. Irma was a Category 5 hurricane and the most powerful hurricane on record in the open Atlantic region outside of the Caribbean Sea and Gulf of Mexico. Irma was also the strongest tropical cyclone by wind speed worldwide in 2017. Irma caused widespread and catastrophic damage throughout its path, and was particularly severe in the northeastern Caribbean. 
        
        Irma developed from a tropical wave near the Cape Verde Islands. Irma then became a Category 3 hurricane on the Saffir-Simpson wind scale before resuming intensifying on September 4 and becoming a Category 5 hurricane by early September 5. Irma's intensity peaked on September 6 with 1-minute sustained winds at 180mph and a minimum pressure of 914 hPa. Before making landfall in Cuba, Irma weakened to a Category 4 hurricane, but regained its Category 5 status before hitting Cuba. 
        
        Irma hit both Caribbean islands and the continental United States. Irma caused catastrophic damage in Barbuda, Saint Barthélemy, Saint Martin, Anguilla, and the Virgin Islands as a Category 5 hurricane. Irma also made landfall in Anguilla, Barbados, Cuba, French West Indies, Haiti, Puerto Rico, and the Dutch side of Sint Maarten. After crossing the Straits of Florida, Irma made landfall in Cudjoe Key on September 10 making landfall in Florida before then making landfall at Marco Island.
        
    \\
    
    \bottomrule
    \end{tabular}
    \caption{Hurricane Irma}
    \label{tab:model_human_ref2}
\end{table*}

\end{document}